\documentclass[]{antgroup}
\PassOptionsToPackage{numbers, compress}{natbib}
\usepackage{antgroup}

\usepackage{graphicx}
\usepackage{tikz}
\usepackage{todonotes}
\usepackage{multirow}
\usepackage{amsmath}
\usepackage{cleveref}
\usepackage{subcaption}
\usepackage{multicol}
\usepackage{amssymb}
\usepackage{array}
\usepackage{bm}
\usepackage{enumitem}
\usepackage{algorithm}
\usepackage{algpseudocode}
\usepackage{tabularx}
\usepackage{booktabs} 
\usepackage{bbm}
\usepackage{makecell}
\usepackage{wrapfig}
\usepackage{colortbl}
\usepackage{threeparttable}
\usepackage{fontawesome5}
\usepackage[normalem]{ulem}

\useunder{\uline}{\ul}{}

\usepackage{amsmath,amsfonts,bm}

\def\eqref#1{equation~\ref{#1}}
\def\1{\bm{1}}

\DeclareMathAlphabet{\mathsfit}{\encodingdefault}{\sfdefault}{m}{sl}
\SetMathAlphabet{\mathsfit}{bold}{\encodingdefault}{\sfdefault}{bx}{n}

\newcommand*\justify{%
  \fontdimen2\font=0.4em% interword space
  \fontdimen3\font=0.2em% interword stretch
  \fontdimen4\font=0.1em% interword shrink
  \fontdimen7\font=0.1em% extra space
  \hyphenchar\font=`\-% allowing hyphenation
}

\renewcommand{\texttt}[1]{%
  \begingroup
  \ttfamily
  \begingroup\lccode`~=`/\lowercase{\endgroup\def~}{/\discretionary{}{}{}}%
  \begingroup\lccode`~=`[\lowercase{\endgroup\def~}{[\discretionary{}{}{}}%
  \begingroup\lccode`~=`.\lowercase{\endgroup\def~}{.\discretionary{}{}{}}%
  \catcode`/=\active\catcode`[=\active\catcode`.=\active
  \justify\scantokens{#1\noexpand}%
  \endgroup
}

\usepackage{amsmath}
\usepackage{amssymb}
\usepackage{amsfonts}                               % blackboard math symbols
\usepackage{amsthm}
\usepackage[mathcal]{eucal}
\usepackage{mathrsfs}
\usepackage{bm}                                     % bm command
\usepackage{blkarray}                               % to support matrix
\usepackage{nicefrac}                               % compact symbols for 1/2, etc.

\usepackage{wrapfig}
\usepackage{graphicx}                               % include pdf figures
\usepackage{caption}
\usepackage{cleveref}
\usepackage{tikz}                                          
\usepackage{circuitikz}
\usetikzlibrary{patterns,snakes}
\usetikzlibrary{positioning,calc,fit,decorations.pathmorphing,shapes.geometric, shapes.gates.logic.US, calc}
\usetikzlibrary{arrows,arrows.meta,decorations.markings,shapes,shapes.arrows}
\usetikzlibrary{decorations,decorations.pathreplacing}
\usetikzlibrary{backgrounds}
\usepackage{filecontents}                           % support to pgfplots
\usepackage{pgfplots}
\usepackage{pgfplotstable}
\usepgfplotslibrary{groupplots}
\usepackage{scalefnt}
\pgfplotsset{compat=newest}
\usepackage{xcolor}

\definecolor{LightGray}{gray}{0.9}

\definecolor{myPink}{HTML}{EE6B98}
\definecolor{myBlue}{HTML}{6BB8FA}
\definecolor{myPurple}{HTML}{9F63F0}
\definecolor{myBlueBase}{HTML}{0070C0}

\title{MultiEdit: Advancing Instruction-based Image Editing\\on Diverse and Challenging Tasks}

\author{Mingsong Li$^{1,2}$, Lin Liu$^{1}$, Hongjun Wang$^{1,3}$, Haoxing Chen$^{1}$, Xijun Gu$^{1,4}$, \\Shizhan Liu$^{1}$, Dong Gong$^{2}$, Junbo Zhao$^{1,4}$, Zhenzhong Lan$^{1,5}$, Jianguo Li$^{1\dag}$}

\footnotetext{$^\dag$Corresponding Authors.}

\affiliation{$^1$Inclusion AI\quad}
\affiliation{$^2$University of New South Wales\\}
\affiliation{$^3$The University of Hong Kong\quad}
\affiliation{$^4$Zhejiang University\quad}
\affiliation{$^5$Westlake University\quad}

\begin{document}
\maketitle

%%%%%%%%%%%%%%%%%%%%%%%%%%%%%%%%%%%%%%%%%%%%%%%%%%%%%%%%%%%%%%%%%%
%%%%%%%%%%%%%%%%%%%%%%%%%%%%%%%%%%%%%%%%%%%%%%%%%%%%%%%%%%%%%%%%%%
%%%%%%%%%%%%%%%%%%%%%%%%%%%%%%%%%%%%%%%%%%%%%%%%%%%%%%%%%%%%%%%%%%
% main document
\begin{abstract}
Current instruction-based image editing (IBIE) methods struggle with challenging editing tasks, as both editing types and sample counts of existing datasets are limited. Moreover, traditional dataset construction often contains noisy image-caption pairs, which may introduce biases and limit model capabilities in complex editing scenarios.
To address these limitations, we introduce \textbf{MultiEdit}, a comprehensive dataset featuring over 107K high-quality image editing samples. It encompasses 6 challenging editing tasks through a diverse collection of 18 non-style-transfer editing types and 38 style transfer operations, covering a spectrum from sophisticated style transfer to complex semantic operations like person reference editing and in-image text editing.
We employ a novel dataset construction pipeline that utilizes two multi-modal large language models (MLLMs) to generate visual-adaptive editing instructions and produce high-fidelity edited images, respectively. 
Extensive experiments demonstrate that fine-tuning foundational open-source models with our MultiEdit-Train set substantially improves models' performance on sophisticated editing tasks in our proposed MultiEdit-Test benchmark, while effectively preserving their capabilities on the standard editing benchmark. 
We believe MultiEdit provides a valuable resource for advancing research into more diverse and challenging IBIE capabilities. Our dataset is available at \url{https://huggingface.co/datasets/inclusionAI/MultiEdit}.
\end{abstract}

\section{Introduction}
Image editing~\citep{zhu2017unpaired,couairon2023diffedit,zhang2023sine,huang2025diffusion} seeks to enable controllable image generation by modifying target components while preserving the integrity of other visual details. Among various editing paradigms, instruction-based image editing (IBIE)~\citep{hui2024hq,Magicbrush,Emuedit,yu2024anyedit,shi2024seededit0,zhang2025context} provides an intuitive and flexible approach, enabling users to modify images simply by providing natural language commands, without the need for complex descriptions or region-specific masks. While IBIE presents a promising avenue, current models still face challenges in consistently executing a wide array of instructions with high fidelity. This difficulty can be primarily attributed to a primary bottleneck: the limitations of available training data.

\begin{figure*}[!t]
\centering
\includegraphics[width=0.975\linewidth]{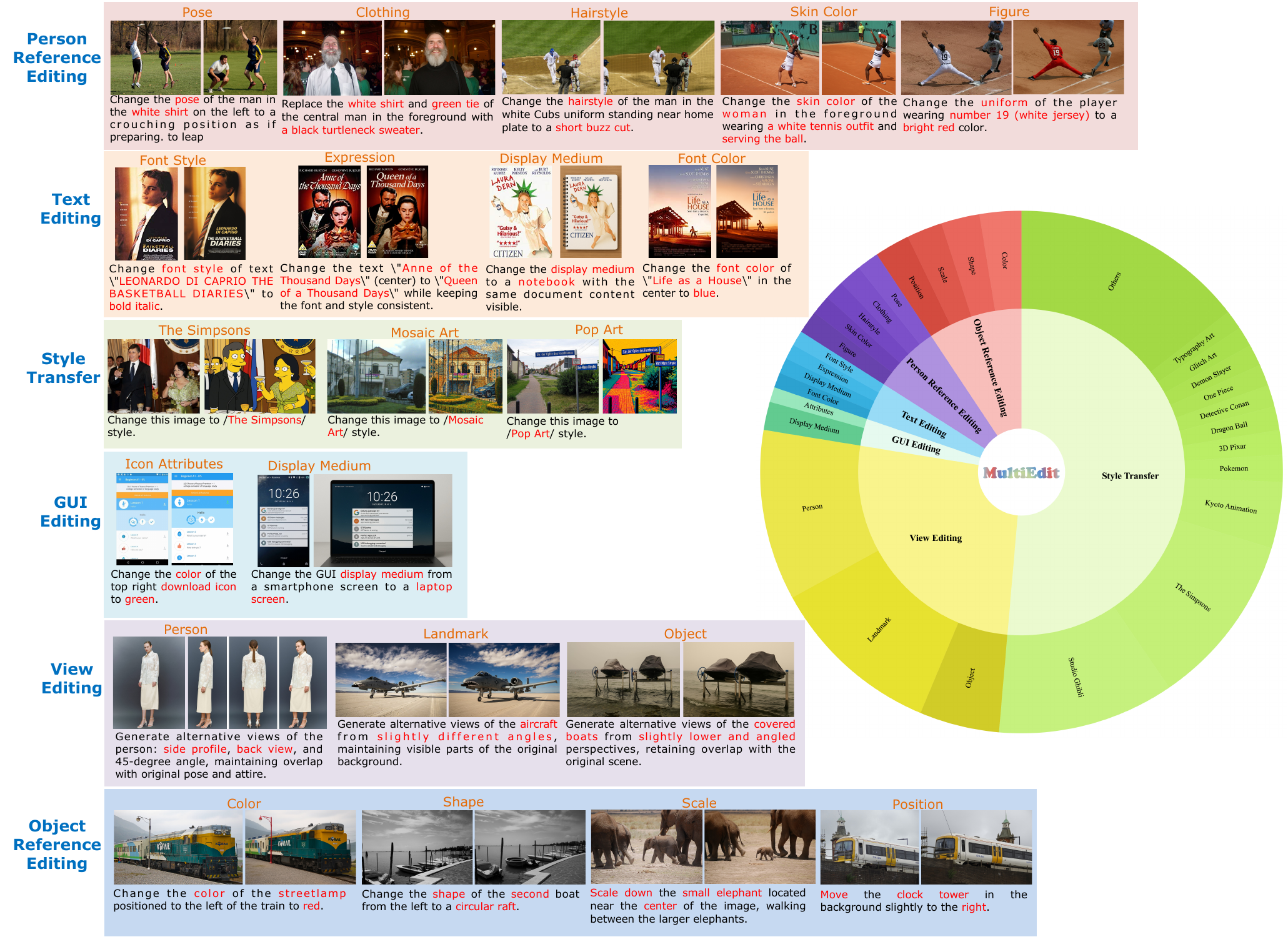}
\caption{Overview of the MultiEdit dataset, illustrating its qualitative diversity and compositional structure. The left panel showcases the qualitative diversity of our 6 editing tasks, featuring representative examples from all 18 non-style-transfer editing types and several of the 38 style transfer operations. On the right, the sunburst chart outlines task distribution within MultiEdit, where the inner ring shows the 6 categories and the outer ring details the distribution of specific editing types. This dual perspective highlights the breadth and depth of challenging image editing scenarios covered in MultiEdit.}
\label{fig:data_distribution}
\end{figure*}

Specifically, existing datasets~\citep{Magicbrush,Emuedit,zhao2024ultraedit,yu2024anyedit} focus primarily on relatively simple edits within natural images, such as local or global modifications of single entities. They largely neglect more complex and fine-grained scenarios essential for real-world applications, such as edits requiring spatial or semantic reasoning (e.g., object reference editing and view editing) and manipulations within structured images (e.g., movie posters and GUI interfaces).

Addressing this critical data gap presents two fundamental challenges. First, conventional data construction pipelines~\citep{Magicbrush,Emuedit,zhao2024ultraedit,yu2024anyedit} are potentially flawed. They typically utilize large language models (LLMs) to rephrase an image's original caption into an editing instruction and a target caption. This approach is problematic because captions, as textual abstractions of rich visual content, are susceptible to information loss, noise, and misalignment~\citep{yu2024anyedit,li2025superedit,wang2025gpt}. These fundamental flaws are not only preserved but often amplified during the rephrasing process, rendering the entire pipeline prone to generating low-quality data or prohibitively labor-intensive to correct. Second, a key hurdle lies in the practical execution of advanced edits, as existing editing models often struggle to achieve high-fidelity results for sophisticated tasks without specific training data. However, the recent introduction of the SOTA ImageGen model (i.e., GPT-Image-1~\citep{gpt-4o,gpt-image-1}) represents a pivotal milestone in multi-modal generative modeling. Its exceptional capacity in complex compositional editing and fine-grained attribute manipulation reveals vast yet largely untapped potential for IBIE tasks.

To address these challenges, we introduce \textbf{MultiEdit}, a large-scale dataset featuring over 107K fine-grained triplets of original images, editing instructions, and edited images. The dataset is meticulously designed to address 6 challenging editing tasks through a comprehensive collection of 56 editing types, spanning 18 distinct non-style-transfer operations and 38 style variations. This composition covers a broad spectrum from creative style transfer to complex semantic operations like reference editing and in-image text editing, with distribution visualized in Fig.~\ref{fig:data_distribution} and statistical details in Tab.~\ref{tab:data_summary}. MultiEdit's construction is guided by a novel multi-modal large language model (MLLM)-driven pipeline with two key innovations. First, to counter potentially flawed conventional pipelines~\citep{Magicbrush,Emuedit,zhao2024ultraedit,yu2024anyedit}, our approach utilizes the SOTA MLLM~\citep{gpt-4o} to directly interpret source images and generate highly relevant, diverse, and adaptive editing instructions—termed \emph{visual-adaptive editing instructions}, which ensures firm visual grounding while circumventing reliance on potentially inaccurate captions. Second, to ensure high-fidelity realization of these advanced edits, we leverage the outstanding generative capabilities of the SOTA ImageGen model (i.e., GPT-Image-1~\citep{gpt-image-1}) to generate target edited images, producing faithful visual outcomes for our fine-grained instructions.

To facilitate comprehensive evaluations of IBIE model performance on these complex tasks, we construct MultiEdit-Test, a carefully curated benchmark comprising 1,100 high-quality samples from MultiEdit, covering all targeted tasks and editing types (50 samples per editing type). The remaining 106,534 samples form the large-scale MultiEdit-Train set for model training. Through experiments in various open source models~\citep{esser2024scaling,zhao2024ultraedit,yu2024anyedit,liu2025step1x}, we demonstrate that fine-tuning foundational models with our MultiEdit-Train set substantially elevates their sophisticated editing abilities while maintaining performance on the standard editing benchmark.

Our main contributions are summarized as follows: 
\begin{itemize}
\item We introduce MultiEdit, a comprehensive large-scale IBIE dataset comprising over 107K samples targeting 6 challenging image editing tasks covering 56 subcategory editing types (18 non-style-transfer and 38 style transfer). We also release MultiEdit-Test, a carefully curated benchmark of 1.1K samples to assess complex editing performance.

\item We design a novel MLLM-driven data construction pipeline that leverages the SOTA MLLM for visual-adaptive instruction generation directly from source images and the SOTA ImageGen for high-fidelity edited image generation, circumventing limitations of caption-based approaches.

\item We conduct extensive experiments demonstrating that fine-tuning open-source models with our MultiEdit-Train set substantially improves model performance on sophisticated editing tasks while maintaining their capabilities on the standard editing benchmark. We also explore various multi-task learning strategies to establish effective fine-tuning practices for our MultiEdit dataset. 
\end{itemize}

\begin{table*}[!t]
    \centering
    \caption{Detailed composition of the MultiEdit dataset, which encompasses 107,634 samples covering diverse editing scenarios from object and person reference editing to creative style transfer tasks. For each of the 6 task categories, we show the source datasets, specific editing types with their corresponding counts, and the precise split of samples into MultiEdit-Train (106,534) and MultiEdit-Test (1,100) sets.
    }
    \renewcommand{\arraystretch}{1.4}
    \setlength{\tabcolsep}{5pt} % Adjusted for clarity and readability
    \resizebox{\textwidth}{!}{
    \begin{tabular}{cc cc cc}
    \toprule
    \multirow{2}{*}{\textbf{Tasks}} & \multirow{2}{*}{ \textbf{Detailed Source Datasets}} & \multirow{2}{*}{\textbf{editing types}}  & \multirow{2}{*}{\textbf{Train Samples}} & \multirow{2}{*}{\textbf{Test Samples}} & \multirow{2}{*}{\textbf{Total Samples}}\\
        & & & &\\
    \hline
    \multirow{2}{*}{Object Reference Editing} & \multirow{2}{*}{MS-COCO-object} & color (2,778), shape (2,565), & \multirow{2}{*}{9,851} & \multirow{2}{*}{200} & \multirow{2}{*}{10,051} \\
    & & scale (2,370), and position (2,338) & & & \\
    \hline
    \multirow{2}{*}{Person Reference Editing} & \multirow{2}{*}{MS-COCO-person} & pose (1,463), clothing (1,526), hairstyle (1,326), & \multirow{2}{*}{6,891} & \multirow{2}{*}{250} & \multirow{2}{*}{7,141} \\
     & & skin color (1,306), and figure (1,520) & & & \\
    \hline
    \multirow{2}{*}{Text Editing} & \multirow{2}{*}{Movie-poster}  & font style (1,012), expression (1,002), & \multirow{2}{*}{3,860} & \multirow{2}{*}{200} & \multirow{2}{*}{4,060} \\
     & & display medium (1,039), and font color (1,007) & & & \\
    \hline
    GUI Editing & GUI-world & icon attributes (997) and display medium (1,883) & 2,780 & 100 & 2,880\\
    \hline
    \multirow{3}{*}{View Editing} & IGPair-person & person & 11,316 & 50 & 11,366 \\
    & Google-landmark-DS-V2 & landmark & 11,469 & 50 & 11,519\\
    & DIS5K & object & 5,270 & 50 & 5,320  \\
    \hline
    \multirow{4}{*}{Style Transfer} & IGPair-person & \multirow{4}{*}{\parbox{5.6cm}{\centering 38 styles across 3 groups: Animation Style, Modern and Digital Art,  and Classical and Traditional Art\textsuperscript{\textdagger}}} & 12,957 & 50 & 13,007\\
    & AM2K &  & 4,913 & 50 & 4,963\\
    & Google-landmark-DS-V2 & & 18,379 & 50 & 18,429\\
    & MS-COCO-object & & 18,848 & 50 & 18,898\\
    \hline
    \textbf{Total Count}&-&-&106,534 & 1,100& 107,634 \\
    \bottomrule
    \end{tabular}
    \vspace{1mm} %
    }
    \begin{flushleft}
    \scriptsize
    \textsuperscript{\textdagger}The 38 styles across 3 groups of Style Transfer task: Animation Style (e.g., `Studio Ghibli', `The Simpsons', and `3D Pixar'), Modern and Digital Art (e.g., `Pop Art', `3D Illustration', and `Isometric LEGO'), and Classical and Traditional Art (e.g., `Mosaic Art', `Paper Cut', and `Xu Beihong').
    \end{flushleft}
    \label{tab:data_summary}
\end{table*}

\section{MultiEdit}
\subsection{Editing Type Definition}
MultiEdit serves as a comprehensive supplement to existing foundational datasets by specifically targeting complex IBIE tasks. As illustrated in Fig.~\ref{fig:data_distribution} and Tab.~\ref{tab:data_summary}, the dataset encompasses 6 specially designed task categories, with each data instance forming a triplet of an original image, an editing instruction, and the corresponding edited image. Spanning these categories, the dataset includes 18 distinct non-style-transfer editing types and a comprehensive style transfer component comprising 38 styles across 3 groups.

The 6 task categories are detailed as follows:

\textbf{(1) Object reference editing} modifies specific attributes (color, shape, scale, and position) of referenced objects, using images from non-person classes of MS-COCO~\citep{lin2014microsoft}.

\textbf{(2) Person reference editing} targets referenced individuals within images, altering their pose, clothing, hairstyle, skin color, and figure, using images from the person class of MS-COCO.

\textbf{(3) Text editing} focuses on textual elements within movie posters, covering modifications in font style, expression, display medium, and font color, using images from the Movie-poster dataset~\citep{dewidar2019inferring}.

\textbf{(4) GUI editing} modifies icon attributes and the display medium of GUI elements, using images of diverse digital interfaces (iOS, Android, and websites) from the GUI-world dataset~\citep{chen2024gui}.

\textbf{(5) View editing} generates alternative views of subjects within images, encompassing edits for persons (from IGPair~\citep{shen2025imagdressing}), landmarks (from Google-landmark-DS-V2~\citep{weyand2020google}), and general objects (from DIS5K~\citep{qin2022highly}).

\textbf{(6) Style transfer} applies new artistic styles to images sourced from IGPair-person, AM2K~\citep{li2022bridging}, Google-landmark-DS-V2, and MS-COCO-object. It comprises 38 distinct styles organized into 3 groups: Animation Styles (e.g., `Studio Ghibli' and `The Simpsons'), Modern and Digital Art (e.g., `Pop Art' and `Geometric Low Poly'), and Classical and Traditional Art (e.g., `Oil Painting' and `Paper Cut').

Representative examples of these editing tasks are shown in Figs.~\ref{fig:data_distribution} and~\ref{fig:style_example}.

\begin{figure*}[!t]
\centering
\includegraphics[width=0.9\linewidth]{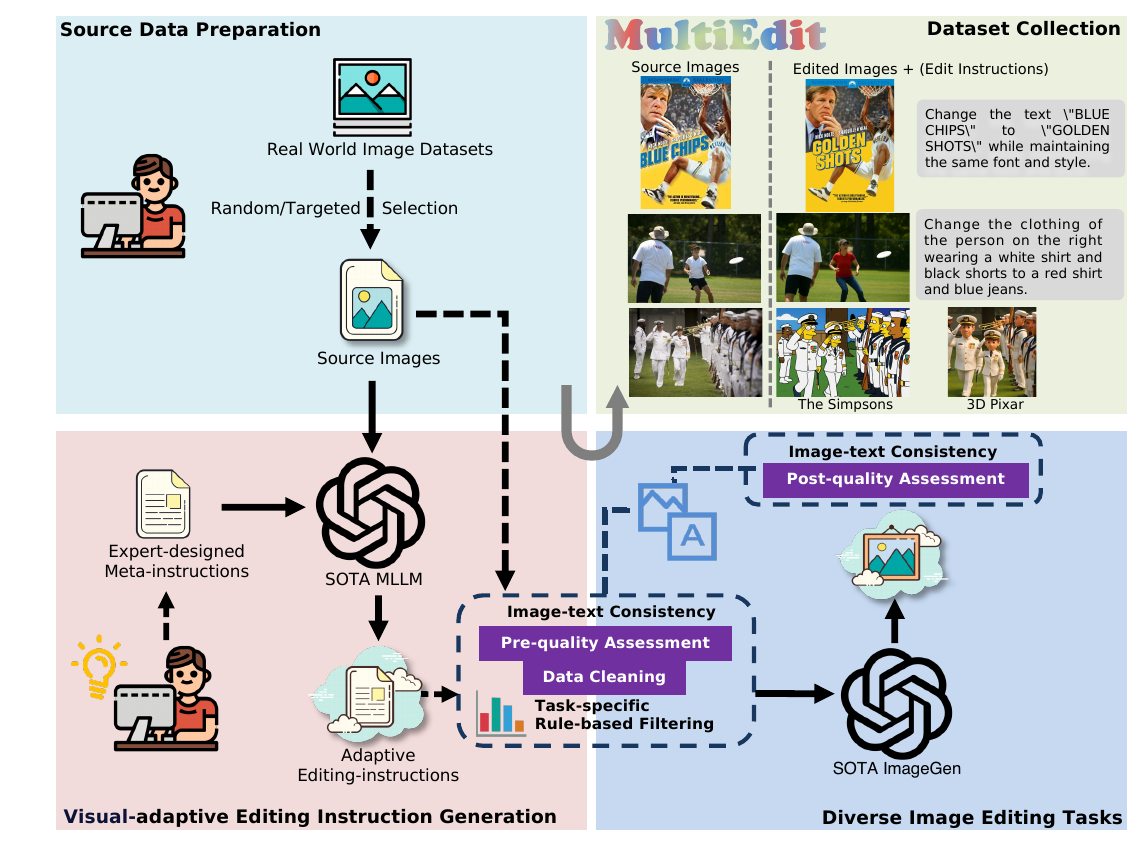}
\caption{Overview of our MLLM-driven data construction pipeline for MultiEdit. The pipeline consists of three main stages with integrated quality checks: (1) Source data preparation through strategic selection from diverse real-world datasets; (2) Visual-adaptive editing instruction generation using the SOTA MLLM, incorporating expert-designed meta-instructions followed by data cleaning and pre-quality assessment; (3) Diverse-task image editing with the SOTA ImageGen, culminating in comprehensive post-quality assessment to ensure dataset quality.}
\label{fig:pipeline}
\end{figure*}

\subsection{Dataset Collection}
To overcome the complexities and misalignments of conventional caption-based methods~\citep{Magicbrush,Emuedit,zhao2024ultraedit,yu2024anyedit}, we introduce a novel MLLM-driven pipeline for MultiEdit. This pipeline operates solely on source images, utilizing the SOTA MLLM for instruction generation and the SOTA ImageGen model for image editing, which significantly streamlines data construction.

\paragraph{Source Data Preparation.} 
As shown in Fig.~\ref{fig:pipeline}, the initial stage involves preparing source images from real-world datasets through two strategies: random and targeted selection. For random selection, applied to the majority of our primary tasks, we randomly sample source images from established public datasets such as Movie-poster and IGPair-person until the pre-defined quotas are met. For targeted selection, employed for specialized tasks, we perform more intricate curation. For instance, we construct MS-COCO-person and MS-COCO-object subsets by sampling exclusively from the `person' and all other categories within the MS-COCO dataset, respectively. Similarly, for GUI editing, we derive source images by first extracting frames from videos in the GUI-world dataset and then performing random sampling. This dual-strategy approach yields a collection of over 68K source images covering our diverse range of editing tasks.

\paragraph{Visual-adaptive Editing Instruction Generation.}\label{sec:dataset_colloct}
Conventional data construction pipelines rely on abstracted captions that are prone to information loss, misalignment, and noise, building on a potentially flawed foundation. To overcome this limitation, we employ the SOTA MLLM~\citep{gpt-4o} to bypass captions and directly generate editing instructions from source images in our MLLM-driven pipeline, as shown in Fig.~\ref{fig:pipeline}. By integrating expert-designed meta-instructions with input images for various editing tasks, the SOTA MLLM utilizes its advanced visual understanding to generate visual-adaptive editing instructions specific to each image. In Appendix~\ref{sec:app_meta_instruc}, we systematically detail the design of expert-designed meta-instructions for generating precise editing instructions across 6 task categories and 56 edit types. Our approach leverages direct visual grounding to significantly reduce information loss and misalignment issues, generating context-aware and accurate instructions for advanced IBIE tasks. This direct visual grounding approach applies to all non-style-transfer tasks. For style transfer, we adopt a template-based strategy, generating 5 instructions for each source image (i.e., \textit{Change this image to /XXXX/ style.}) based on a prioritized list of styles. Finally, this stage yields a total of 136K source image and editing instruction pairs. 

\paragraph{Diverse Image Editing Tasks.}
While our pipeline excels at generating fine-grained visual-adaptive editing instructions, a key hurdle lies in their practical execution. Due to the lack of specific training data for our targeted complex scenarios, existing open-source models often struggle to render such sophisticated edits with high fidelity, creating a critical gap between our fine-grained instructions and the resulting visual outcomes. To bridge this gap, our MLLM-driven framework employs the powerful multi-modal generative capabilities of the SOTA ImageGen~\citep{gpt-image-1} for high-fidelity image editing. This approach ensures that the generated outcomes not only align with the complexity of diverse and challenging IBIE scenarios but also faithfully adhere to the fine-grained details of each visual-adaptive instruction.

\begin{figure*}[!t]
\centering
\includegraphics[width=0.975\linewidth]{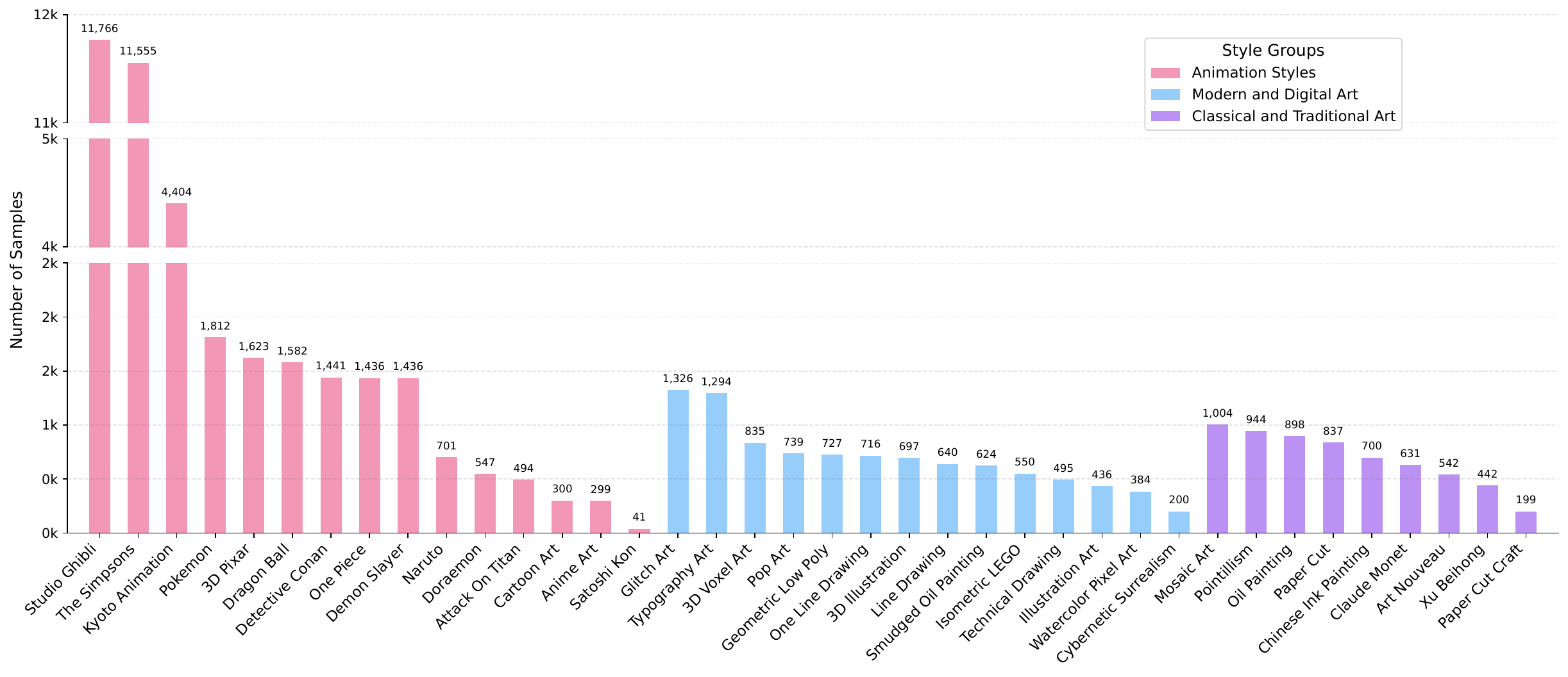}
\caption{Distribution of the style transfer task in MultiEdit. The dataset comprises 55,297 samples across 38 distinct artistic styles, organized into 3 categories: Animation Styles, Modern and Digital Art, and Classical and Traditional Art. With specific sample counts detailed for each style, the distribution strategically emphasizes popular styles like `Studio Ghibli' and `The Simpsons' while maintaining balanced coverage across diverse artistic domains. The broken y-axis accommodates the scale differences.}
\label{fig:style_dist}
\end{figure*}

\paragraph{Data Cleaning and Quality Assessment.}
While our MLLM-driven pipeline is highly automated, achieving superior data quality hinges on validating its two critical outputs, i.e., the generated editing instruction and the edited image. An ideal instruction should be explicit and logically sound, and the corresponding edited image should be faithful to the instruction and free of artifacts. To enforce these standards and ensure overall data quality, as shown in Fig.~\ref{fig:pipeline}, we implement a two-step quality assessment process. 

\textit{(1) Data Cleaning and Pre-quality Assessment.}
This initial step aims to purify the instruction-image pairs through a meticulous process of data cleaning and pre-assessment. The process commences with data cleaning, where we first employ rule-based filters to automatically reject instructions generated from a misalignment between the source image content and the editing type. This step is followed by task-specific curation, such as removing data pairs with low-quality source images (e.g., blurry video frames in GUI editing data). Subsequently, the pre-quality assessment involves evaluating the semantic alignment, clarity, and contextual relevance between each source image and its corresponding instruction.

\textit{(2) Post-quality Assessment.}
Following diverse-task image editing, the post-quality assessment critically evaluates the generated images based on two key criteria: fidelity, ensuring the consistency of edited regions with the corresponding editing instructions, and preservation, maintaining the integrity of non-edited regions. To efficiently execute this assessment, we employ a strategic two-tiered review process. Initially, a small set of samples from each generation task undergoes a manual accuracy check. Tasks falling below a 70\% accuracy threshold are flagged for a comprehensive manual review of all their images. Conversely, for tasks exceeding the 70\% threshold, we leverage Qwen2.5-VL~\citep{bai2025qwen2.5} for assisted screening, with human experts providing oversight on ambiguous cases to guarantee final quality.

\subsection{Characteristics and Statistics}
Constructed through our streamlined pipeline, MultiEdit ultimately comprises over 107K high-quality, fine-grained triplets, each consisting of an original source image, a corresponding visual-adaptive editing instruction, and the resultant edited image. The scope of the involved 6 editing tasks within MultiEdit is notably broad and advanced, spanning from complex semantic operations like reference-based object or person editing and in-image text editing, to specialized applications such as GUI editing and generative view editing for diverse subjects, including persons, landmarks, and objects. 

\begin{figure*}[!t]
\centering
\includegraphics[width=0.975\linewidth]{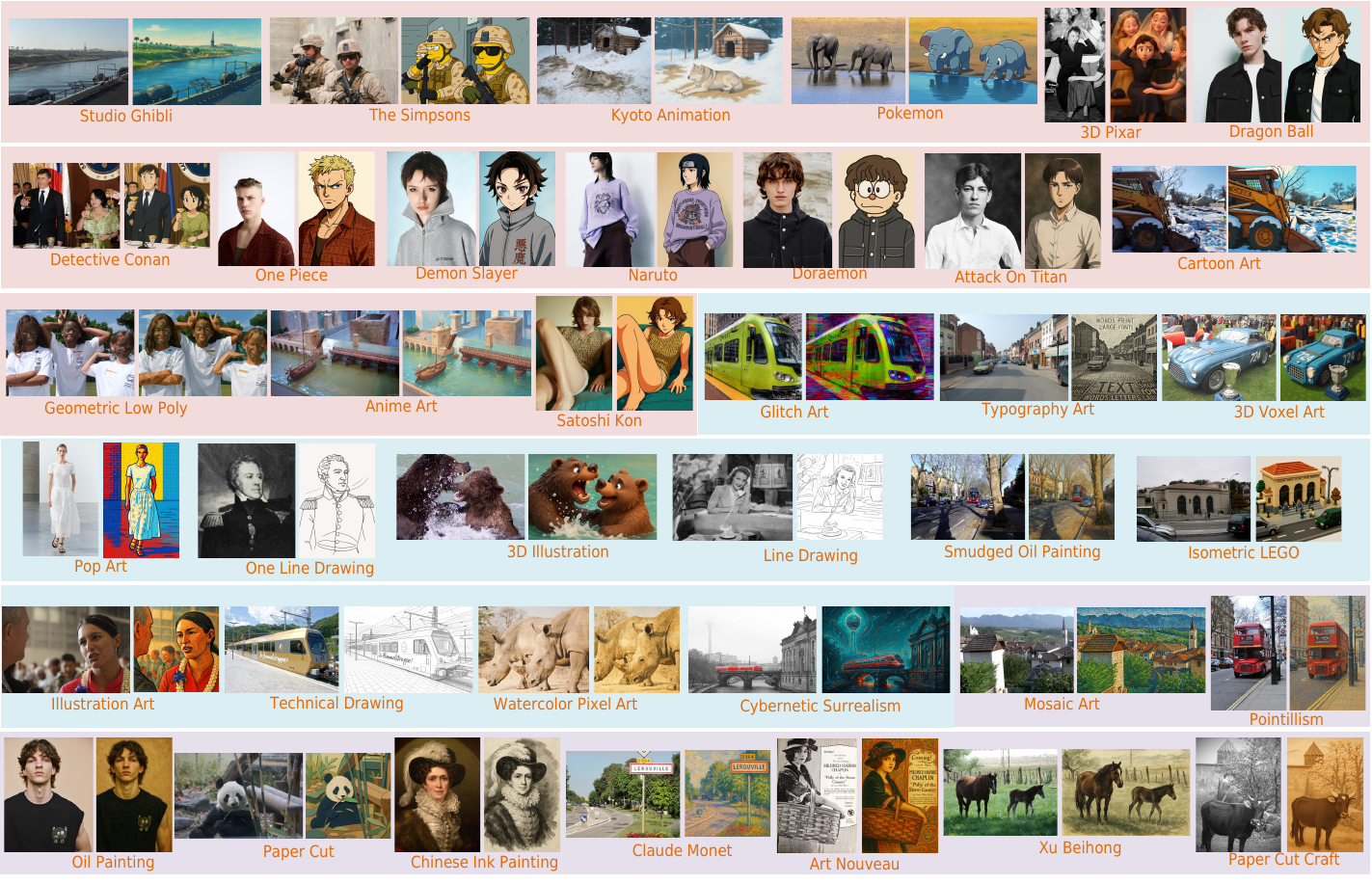}
\caption{Representative examples of all 38 styles from the style transfer subset of MultiEdit, grouped by category. Background colors correspond to the 3 groups: Animation Styles (\textcolor{myPink}{pink}), Modern and Digital Art (\textcolor{myBlue}{blue}), and Classical and Traditional Art (\textcolor{myPurple}{purple}). Best viewed when zoomed in.}
\label{fig:style_example}
\end{figure*}

A visual overview of MultiEdit is provided in Fig.~\ref{fig:data_distribution}, which showcases representative examples with a sunburst chart illustrating the compositional structure. For a detailed breakdown, Tab.~\ref{tab:data_summary} presents the precise composition for each task, including source datasets, precise train-test splits, and the sample distribution across specific editing types. Beyond these core tasks, MultiEdit features a dedicated style transfer component, comprising 55,297 samples across 38 distinct styles organized into 3 groups. Fig.~\ref{fig:style_dist} quantifies the precise sample count for each of all 38 styles, revealing a distribution strategically weighted towards highly popular styles like `Studio Ghibli' and `The Simpsons' while maintaining broad coverage of others. Complementing this distributional view, Fig.~\ref{fig:style_example} provides a qualitative showcase of all 38 styles, each illustrated with a representative source-edit pair.

\subsection{Multi-task Learning Framework}\label{sec:dataset_MTL}
The heterogeneous nature of MultiEdit, encompassing 6 distinct task categories with varying complexity and data distributions, naturally forms as a multi-task learning (MTL) problem~\citep{kendall2018multi, liu2019end, yu2020gradient, liu2021conflict}. Specifically, our dataset exhibits a hierarchical complexity structure: style transfer tasks are data-abundant yet relatively straightforward, tasks such as object and person reference editing and view editing present moderate complexity with semantic reasoning requirements, while text and GUI editing tasks constitute the most challenging scenarios with related fewer training samples. This inherent task imbalance and complexity heterogeneity necessitate a systematic MTL approach to effectively leverage the diverse supervision signals across all editing tasks.

Within the MTL framework, existing optimization strategies can be systematically categorized across three complementary dimensions. 
\begin{itemize}
\item \textbf{Data-driven MTL (DMTL) solutions} address task imbalance through sophisticated sampling strategies, including task-weighted data sampling, curriculum learning schedules~\citep{bengio2009curriculum}, and adaptive balancing techniques that dynamically adjust training data distribution based on task difficulty and learning progress. In our work, we explore a data-driven curriculum by partitioning our MultiEdit-Train set $\mathcal{D}_{\text{train}}$ into $K$ complexity-based task groups $\{\mathcal{D}_k\}$. During each training stage $s$, a weighted random sampler determines the selection probability for any sample $x_i \in \mathcal{D}_k$ based on its group's pre-defined weight $w_k^{(s)}$:
\begin{align} \label{eq:dmtl}
P(x_i \mid \mathcal{D}_k, s) = \frac{w_k^{(s)}}{\sum_{j=1}^{K} |\mathcal{D}_j| \cdot w_j^{(s)}}.
\end{align}
By adjusting the weight vector $\mathbf{w}^{(s)}$ between stages, we can dynamically shift the training focus among the pre-defined task groups as training progresses.

\item \textbf{Architecture-driven MTL (AMTL) solutions} tackle the challenge of parameter sharing by designing specialized network components, ranging from task-specific pathways~\citep{misra2016cross,chen2023mod} and cross-task attention mechanisms~\citep{liu2019end} to adaptive parameter allocation strategies that optimize capacity distribution across different learning objectives.

\item \textbf{Loss-driven MTL (LMTL) solutions} focus on gradient dynamics and objective balance, encompassing uncertainty-based loss weighting~\citep{kendall2018multi}, gradient conflict mitigation techniques~\citep{yu2020gradient}, and adaptive loss scaling methods that ensure balanced learning across heterogeneous tasks. In our work, we explore two sample-wise strategies for re-weighting the per-sample loss $\ell_i$, computed as:
\begin{align}
    \ell_i = \mathrm{Mean}\left[\, \omega_t \cdot (\hat{\epsilon}_i - \epsilon_i)^2 \,\right],
\end{align}
where $\hat{\epsilon}_i$ is the model's noise prediction, $\epsilon_i$ is the target noise, $\omega_t$ is the timestep-dependent weighting factor, and $\mathrm{Mean}[\cdot]$ denotes spatial averaging over latent dimensions. The final re-weighted loss $\mathcal{L}_i$ for sample $i$ is then given by:
\begin{align}
    \mathcal{L}_i = \omega_i \cdot \ell_i,
\end{align}
where $\omega_i$ is the sample-wise re-weighting coefficient determined by one of the following strategies.

(1) A heuristic approach sets the weight $\omega_i$ for a sample from task group $k$ within a mini-batch $\mathcal{B}$ based on its inverse frequency with a smoothing factor $\alpha$:
\begin{align}\label{eq:lmtl_task_group}
    \omega_i \propto (N_k(\mathcal{B}) + \alpha)^{-1},
\end{align}
where $N_k(\mathcal{B})$ denotes the count of samples from group $k$ in the current mini-batch $\mathcal{B}$. 

(2) A dynamic approach derives the weight $\omega_i$ from the $L_{2}$ norm of the gradient of the per-sample loss with respect to the model's noise prediction, as $G_i = \| \nabla_{\hat{\epsilon}_i} \ell_i \|_2$. We test two weighting schemes: setting $\omega_i$ to be either directly proportional to $G_i$ to amplify harder examples, or inversely proportional to $G_i$ to stabilize training:
\begin{align}\label{eq:lmtl_gradient_norm}
    \omega_i \propto G_i \quad (\text{proportional}) \quad \text{or} \quad \omega_i \propto G_i^{-1} \quad (\text{inverse}).
\end{align}

\end{itemize}

\section{Experiments}

\subsection{Experimental Setup}
\noindent
\paragraph{Baselines.}
We conducted experiments with open-source IBIE methods, including Stable Diffusion 3 (SD3)\footnote{SD3: \url{https://huggingface.co/stabilityai/stable-diffusion-3-medium}}~\citep{esser2024scaling}, UltraEdit\footnote{UltraEdit: \url{https://github.com/HaozheZhao/UltraEdit}}~\citep{zhao2024ultraedit}, AnyEdit\footnote{AnyEdit: \url{https://dcd-anyedit.github.io/}}~\citep{yu2024anyedit}, and Step1X-Edit\footnote{Step1X-Edit: \url{https://step1x-edit.github.io/}}~\citep{liu2025step1x}. For SD3 and UltraEdit, we adopt SD3-Medium-based models and use the stage-1 UltraEdit for free-form image editing. For Step1X-Edit, we employ version 1.0.

\paragraph{Benchmarks \& Metrics.}
In addition to our MultiEdit-Test benchmark, we utilize EmuEdit-Test~\citep{Emuedit} to conduct more comprehensive experiments. Following prior work~\citep{zhao2024ultraedit,yu2024anyedit}, we adopt semantic similarity (i.e., CLIP\textsubscript{img} and CLIP\textsubscript{txt}~\citep{radford2021learning}), visual similarity (i.e., DINO~\citep{caron2021emerging}), and pixel-level differences (i.e., L\textsubscript{1} and L\textsubscript{2} distances) as metrics to evaluate the effectiveness of different methods on IBIE tasks.

\paragraph{Implementation Details.}
To ensure consistent referencing across experiments, we adopt the following notation to denote all model variants trained on our MultiEdit-Train (ME) dataset: ME-\{Base-Model\}-\{Training-Method\}-\{Dataset-Info\}, and the components are:
\begin{itemize}
    \item Base-Model: Base model (SD = SD3-based and UEdit = UltraEdit-based).
    \item Training-Method: Training strategy (SFT = Supervised Fine-tuning, DMTL = Data-driven MTL, and LMTL = Loss-driven MTL).
    \item Dataset-Info: Optional data composition (Mix = MultiEdit-Train + Mixed training data from UltraEdit).
\end{itemize}

For the training setup, we fine-tune models for 5 epochs using the AdamW optimizer~\citep{Loshchilov2017DecoupledWD}. Training uses a constant-with-warm-up learning-rate schedule with final linear annealing, starting at 5e-5 with a 12-step warm-up and linearly decaying to 5e-7 in the final epoch. Across 8 GPUs, we maintain an effective batch size of 256, configured with a per-GPU batch size of 4 and 8 gradient accumulation steps. For mixed data training, warm-up steps are scaled proportionally to the mixed data size (e.g., 24 steps for 100K mixed samples). The detailed model inference guidance experiment and the guidance summary are available in Appendix~\ref{sec:app_guidance}.

\begin{table*}[!t]
\centering
\caption{Quantitative comparison of different methods on MultiEdit-Test and EmuEdit-Test. Rows in gray indicate fine-tuned models with our MultiEdit-Train set. The results demonstrate that foundational models can be significantly improved for our complex editing scenarios by supplementing them with targeted data. Results of AnyEdit on EmuEdit-Test are from its original paper. Best and second-best results are shown in \textbf{bold} and \underline{underlined}, respectively.}
\renewcommand{\arraystretch}{1.25}
\setlength{\tabcolsep}{3pt} 
\footnotesize
\begin{tabular}{lccccccccc}
\toprule
\multirow{2}{*}{\textbf{Method}} & \multicolumn{4}{c}{\textbf{MultiEdit-Test}} & \multicolumn{5}{c}{\textbf{EmuEdit-Test}}\\
\cmidrule(lr){2-5} \cmidrule(lr){6-10}
& CLIP\textsubscript{img}$\uparrow$ & DINO$\uparrow$ & L\textsubscript{1}$\downarrow$ & L\textsubscript{2}$\downarrow$ 
& CLIP\textsubscript{img}$\uparrow$ & CLIP\textsubscript{txt}$\uparrow$ & DINO$\uparrow$ & L\textsubscript{1}$\downarrow$ & L\textsubscript{2}$\downarrow$ \\
\midrule
SD3 & 0.6817 & 0.5751 & 0.1983 & 0.0854  & 0.7019 & 0.2443 & 0.5094 & 0.1708 & 0.0625 \\
UltraEdit &0.8017 &0.7303 &0.1976 &0.0871   & \underline{0.8685} & 0.2706 & \underline{0.8055} & \underline{0.1137} & \bfseries0.0406 \\
AnyEdit & 0.8067 & 0.7246 & 0.1926 & 0.0826 & \bfseries0.8720 & \bfseries0.2850 & \bfseries0.8210 & \bfseries0.0700 & - \\
Step1X-Edit & \bfseries0.8335 & 0.7466 & 0.1954 & 0.0852 & - & - & - & - & - \\
\midrule
\rowcolor{LightGray}
ME-SD-SFT& 0.7759 & 0.7359 & 0.1901 & 0.0858       & 0.7832 & 0.2662 & 0.6431 & 0.2055 & 0.0878 \\
\rowcolor{LightGray}
ME-SD-SFT-Mix100K & 0.7863 & 0.7586 & 0.1926 & 0.0883   & 0.8319 & \underline{0.2723} & 0.7551 & 0.1535 & \underline{0.0619} \\
\rowcolor{LightGray}
ME-UEdit-SFT & 0.8127 & \underline{0.8019} & \underline{0.1827} & \underline{0.0813}     & 0.8384 & 0.2690  & 0.7597 & 0.1903 & 0.0796 \\
\rowcolor{LightGray}
ME-UEdit-DMTL & \underline{0.8174} & \bfseries0.8071 & \bfseries0.1814 & \bfseries0.0800    & 0.8376 & 0.2681 & 0.7610  & 0.1896 & 0.0793 \\
\bottomrule
\end{tabular}
\label{tab:quantitative_results}
\end{table*}

\subsection{Quantitative Evaluation}
We first benchmark four baseline methods on MultiEdit-Test set, evaluating their performance across the diverse tasks proposed in MultiEdit. At the same time, to explore the efficacy of our proposed MultiEdit dataset, we fine-tune two foundational open-source models, SD3 and UltraEdit, on our MultiEdit-Train set. For SD3 and UltraEdit, we only activate MMDiT and keep the rest of the models frozen. To further refine their distinct strengths, we enhance the foundational SD3 via mixed data training with 100K image editing data from UltraEdit. For the more advanced UltraEdit, we adopt the best explored MTL strategy, i.e., the task-weighted data sampling strategy of the data level to modulate the task sampling distribution within our MultiEdit-Train set. Further experiments of these strategies are detailed in Sec.~\ref{sec:dis}.

As illustrated in Tab.~\ref{tab:quantitative_results}, even advanced methods like AnyEdit and Step1X-Edit lack full proficiency in handling the diverse and complex tasks in MultiEdit-Test. This is particularly evident in DINO scores, where Step1X-Edit only reaches approximately 0.75. In contrast, both SD3 and UltraEdit exhibit evident performance improvements on our MultiEdit-Test set after direct SFT. For the SD3 model, this process yields substantial gains on MultiEdit-Test, with CLIP\textsubscript{img} and DINO scores increasing by approximately 9.4\% and 16.1\%, respectively. Notably, this training also boosts its CLIP\textsubscript{img} performance on EmuEdit-Test set by about 8.1\%. Building upon this, our mixed data training strategy provides a further performance lift on both datasets, ultimately achieving peak CLIP\textsubscript{img} and DINO scores of 0.8319 and 0.7551 on EmuEdit-Test. For the UltraEdit model, direct SFT yields a performance increase of approximately 1.1\% in CLIP\textsubscript{img} and 7.2\% in DINO on MultiEdit-Test, while effectively preserving its capabilities on EmuEdit-Test. Adopting the task-weighted data sampling strategy in DMTL further improves its performance, achieving a CLIP\textsubscript{img} score close to the SOTA model Step1X-Edit and surpassing it on DINO by over 5\%. This demonstrates that by supplementing with targeted data, even foundational models can be significantly improved for complex editing scenarios. We also report detailed results of the selected methods across all 6 tasks of our MultiEdit-Test in Appendix~\ref{sec:app_detail_multitest}.

\begin{figure*}[!t]
\centering
\includegraphics[width=0.95\linewidth]{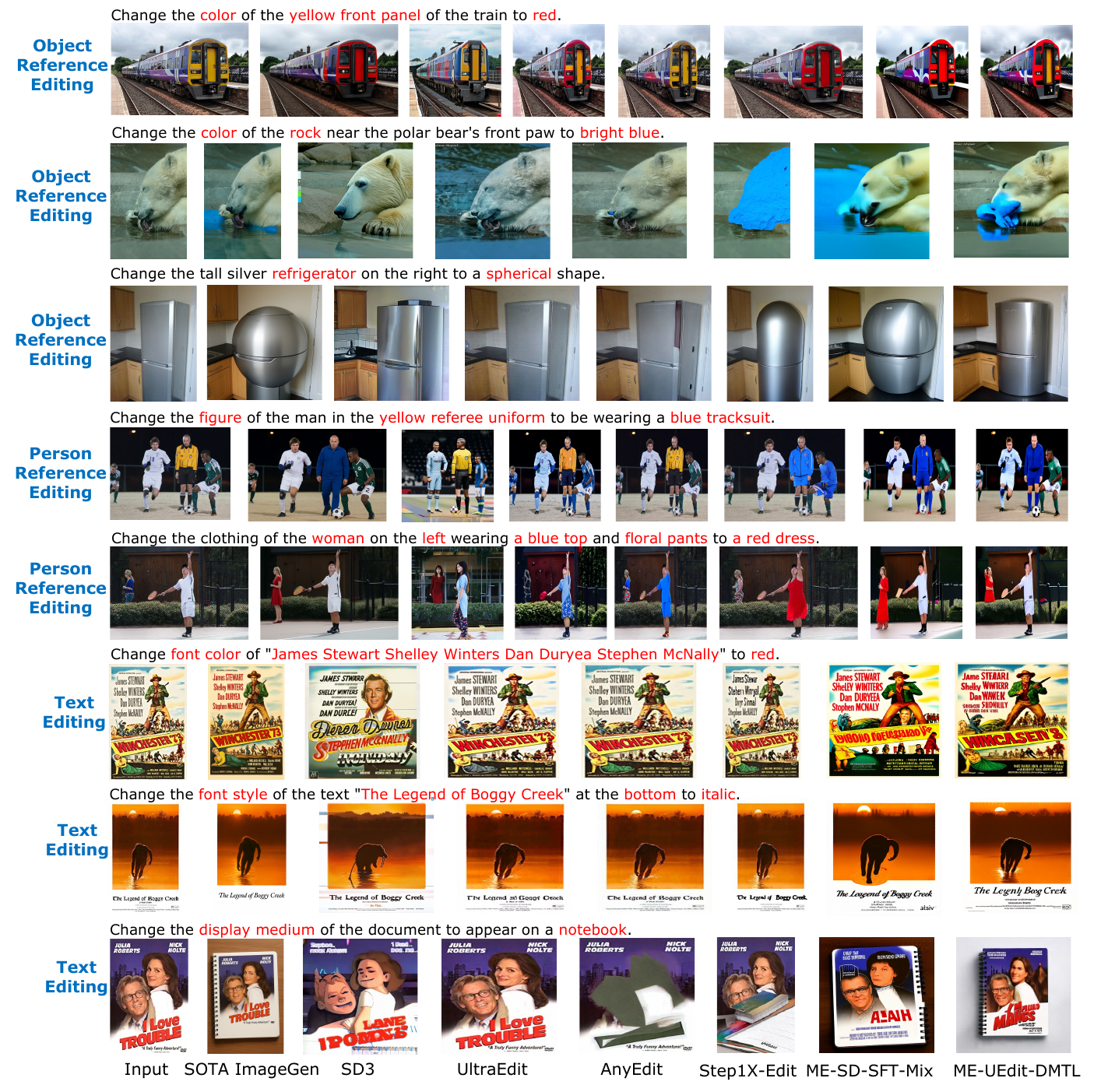}
\caption{Qualitative comparison of different methods across various tasks on MultiEdit-Test, including object reference editing, person reference editing, and text editing. Best viewed when zoomed in.}
\label{fig:qualita_1}
\end{figure*}

\begin{figure*}[!t]
\centering
\includegraphics[width=0.95\linewidth]{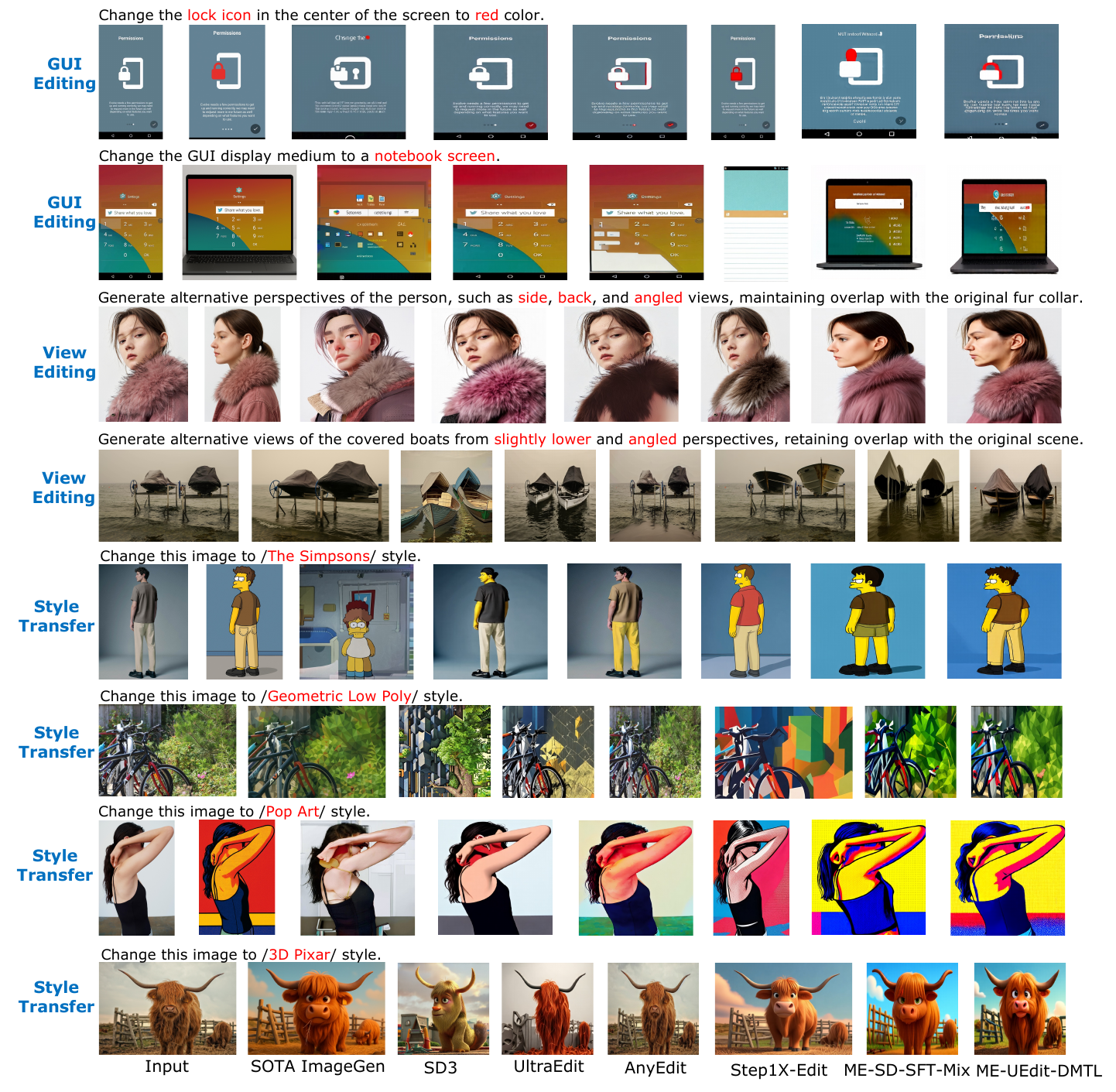}
\caption{Qualitative comparison of different methods across various tasks on MultiEdit-Test, including GUI editing, view editing, and style transfer. Best viewed when zoomed in.}
\label{fig:qualita_2}
\end{figure*}

\subsection{Qualitative Evaluation}
Figs.~\ref{fig:qualita_1} and~\ref{fig:qualita_2} present a qualitative comparison of four baseline methods against our two best-performing fine-tuned models in the 6 MultiEdit-Test task scenarios. From this comparison, we draw several key observations. (1) In line with our quantitative results, fine-tuning with targeted data significantly enhances foundational models like SD3 and UltraEdit for diverse and complex editing tasks, evident in their markedly improved instruction comprehension and precise editing. (2) Despite these improvements, there remains scope for further enhancement for our fine-tuned models regarding fine-grained details, particularly in facial fidelity. (3) The challenges posed by our diverse and complex data scenarios are evident even in the performance of state-of-the-art models, with Step1X-Edit showing limited instruction comprehension in tasks like object reference editing (Fig.~\ref{fig:qualita_1}, row 2) and person reference editing (Fig.~\ref{fig:qualita_1}, row 5). (4) AnyEdit struggles with style transfer, leaving ample room for improvement with the targeted data.

\newtcolorbox{mydisplaybox_0}[1][]{
  enhanced,
  arc=3mm,
  boxrule=0.5pt,
  colback=myBlueBase!10!white,
  colframe=myBlueBase!80!black,
  sharp corners=south,
  fonttitle=\bfseries,
  coltitle=white, 
  attach boxed title to top left={xshift=2mm, yshift=-2mm},
  boxed title style={
    arc=3mm,
    boxrule=0.3pt, 
    colback=myBlueBase!90!white,
    colframe=myBlueBase!80!black, 
    width=\dimexpr\linewidth-2*0.5pt-2*4mm\relax*0.7, % 
  },
  drop shadow={black!20!white},
  #1
}

\section{Discussion}\label{sec:dis}
In this section, we discuss several key strategies for optimizing model training on our MultiEdit-Train set. We primarily investigate multi-task learning (MTL) approaches at the data and loss levels. We also explore the synergy achieved by mixing our MultiEdit-Train set with external data for broader task compatibility.

\begin{table*}[!t]
\centering
\caption{Comparison of data-driven and loss-driven strategies for multi-task fine-tuning on our MultiEdit-Train set. Best results per group are in \textbf{bold}.}
\renewcommand{\arraystretch}{1.25}
\setlength{\tabcolsep}{3pt}
\resizebox{\textwidth}{!}{
\begin{tabular}{lccccccccc}
\toprule
\multirow{2}{*}{\textbf{UltraEdit Config}} & \multicolumn{4}{c}{\textbf{MultiEdit-Test}} & \multicolumn{5}{c}{\textbf{EmuEdit-Test}}\\
\cmidrule(lr){2-5} \cmidrule(lr){6-10}
& CLIP\textsubscript{img}$\uparrow$ & DINO$\uparrow$ & L\textsubscript{1}$\downarrow$ & L\textsubscript{2}$\downarrow$ 
& CLIP\textsubscript{img}$\uparrow$ & CLIP\textsubscript{txt}$\uparrow$ & DINO$\uparrow$ & L\textsubscript{1}$\downarrow$ & L\textsubscript{2}$\downarrow$ \\
\midrule
UltraEdit &0.8017 &0.7303 &0.1976 &0.0871   & 0.8685 & 0.2706 & 0.8055 & 0.1137 & 0.0406 \\
ME-UEdit-SFT & 0.8127 & 0.8019 & 0.1827 & 0.0813     & 0.8384 & 0.2690  & 0.7597 & 0.1903 & 0.0796 \\
\midrule
\multicolumn{10}{l}{\textit{\textbf{ME-UEdit-DMTL:} task-weighted data sampling}} \\
+Setting1\textsuperscript{\textdagger} (1,1,1,5)-(1,1,1,10) & 0.8154 & 0.8038 & 0.1850 & 0.0816 & 0.8326 & \bfseries0.2686 & 0.7494 & 0.1988 & 0.0849 \\
+Setting2 (1,1,1,3.5)-(1,1,1,7.5) & 0.8169 & 0.8056 & 0.1872 & 0.0834 & 0.8337 & \bfseries0.2686 & 0.7488 & 0.2045 & 0.0892 \\
+Setting3 (1.5,1,2,10)-(1.5,1,2.5,15)& 0.8053 & 0.7930 & 0.1828 & 0.0822 & 0.8356 & 0.2681 & 0.7561 & 0.1956 & 0.0833 \\
+Setting4 (1.5,1,2,10)-(1.5,1,3.5,20)& \bfseries0.8174 & \bfseries0.8071 & \bfseries0.1814 & \bfseries0.0800 & \bfseries0.8376 & 0.2681 & \bfseries0.7610 & \bfseries0.1896 & \bfseries0.0793 \\
\midrule
\multicolumn{10}{l}{\textit{\textbf{ME-UEdit-LMTL:} sample-wise loss re-weighting}} \\
+Task Group Balancing ($\alpha=0$) & 0.8118 & 0.7988 & 0.1874 & 0.0841 & 0.8350 & 0.2686 & 0.7538 & 0.1930 & 0.0818 \\
+Task Group Balancing ($\alpha=1.0$) & 0.8100 & 0.7991 & 0.1865 & 0.0832 & 0.8347 & \bfseries0.2689 & 0.7551 & 0.1922 & 0.0809 \\
Gradient Focusing, proportional & 0.8080 & 0.7952 & 0.1856 & 0.0828 & 0.8341 & 0.2679 & 0.7516 & 0.1937 & 0.0819 \\
Gradient Balancing, inverse & \bfseries0.8133 & \bfseries0.8051 & \bfseries0.1834 & \bfseries0.0814 & \bfseries0.8357 & 0.2681 & \bfseries0.7579 & \bfseries0.1910 & \bfseries0.0807 \\
\bottomrule
\end{tabular}
}
\begin{flushleft}
    \scriptsize
    \textsuperscript{\textdagger} Please refer Sec.~\ref{sec:dis_dmtl} for the detailed setting instance of task-weighted data sampling.
    \end{flushleft}
\label{tab:multitask}
\end{table*}

\begin{table*}[!t]
\centering
\caption{Comparison of fine-tuning SD3 with different mixed UltraEdit data. Best results are shown in \textbf{bold}.}
\renewcommand{\arraystretch}{1.25}
\setlength{\tabcolsep}{3pt}
\footnotesize
\begin{tabular}{lccccccccc}
\toprule
\multirow{2}{*}{\textbf{SD3 Config}} & \multicolumn{4}{c}{\textbf{MultiEdit-Test}} & \multicolumn{5}{c}{\textbf{EmuEdit-Test}}\\
\cmidrule(lr){2-5} \cmidrule(lr){6-10}
& CLIP\textsubscript{img}$\uparrow$ & DINO$\uparrow$ & L\textsubscript{1}$\downarrow$ & L\textsubscript{2}$\downarrow$ 
& CLIP\textsubscript{img}$\uparrow$ & CLIP\textsubscript{txt}$\uparrow$ & DINO$\uparrow$ & L\textsubscript{1}$\downarrow$ & L\textsubscript{2}$\downarrow$ \\
\midrule
SD3 & 0.6817 & 0.5751 & 0.1983 & 0.0854  & 0.7019 & 0.2443 & 0.5094 & 0.1708 & 0.0625 \\
ME-SD-SFT & 0.7759 & 0.7359 & 0.1901 & 0.0858       & 0.7832 & 0.2662 & 0.6431 & 0.2055 & 0.0878 \\
\midrule 
+Mix50K & 0.7839 & 0.7554 & \bfseries0.1885 & 0.0857 & 0.8292 & 0.2725 & 0.7435 & 0.1516 & 0.0600 \\
+Mix100K & 0.7863 & 0.7586 & 0.1926 & 0.0883 & 0.8319 & 0.2723  & 0.7551 & 0.1535 & 0.0619 \\
+Mix150K & 0.7856 & 0.7574 & 0.1876 & \bfseries0.0848 & \bfseries0.8409 & 0.2720 & \bfseries0.7668 & \bfseries0.1464 & \bfseries0.0578 \\
+Mix200K & \bfseries0.7876 & \bfseries0.7611 & 0.1972 & 0.0933 & 0.8376 & \bfseries0.2729 & 0.7573 & 0.1572 & 0.0645 \\
\bottomrule
\end{tabular}
\label{tab:sd3_mix}
\end{table*}

\subsection{Multi-task Optimization Strategies}\label{sec:dis_multi_task}

\paragraph{Task-weighted Data Sampling.}\label{sec:dis_mixed_data}
To validate the data-driven MTL (DMTL) strategy formulated in Sec.~\ref{sec:dataset_MTL}, we implement a heuristic task-weighted data sampling scheme guided by pre-defined complexity groups. Specifically, we categorize our tasks into four groups: G1 (style transfer), G2 (object reference editing and view editing (object)), G3 (person reference editing, view editing (person), and view editing (landmark)), and G4 (text editing and GUI editing). Following the curriculum learning principle~\citep{bengio2009curriculum}, we apply a two-stage training regimen with distinct sampling weight vectors $\mathbf{w}^{(s)}$ (initial 2 epochs and subsequent 3 epochs). The specific weighting configuration for Setting1 is as follows:\label{sec:dis_dmtl}

\begin{mydisplaybox_0}[title={Task-weighted data sampling configuration instance: Setting1}]
Stage 1 (epochs 0-1): 
weight vector $\mathbf{w}^{(1)} = (w_{G1}^{(1)}, w_{G2}^{(1)}, w_{G3}^{(1)}, w_{G4}^{(1)}) = (1, 1, 1, 5)$; \\
Stage 2 (epochs 2-4): 
weight vector $\mathbf{w}^{(2)} = (w_{G1}^{(2)}, w_{G2}^{(2)}, w_{G3}^{(2)}, w_{G4}^{(2)}) = (1, 1, 1, 10)$
\end{mydisplaybox_0}

As quantified in the DMTL part of Tab.~\ref{tab:multitask}, while standard SFT improves upon the original UltraEdit, our task-weighted data sampling strategies yield further performance gains. Notably, Setting4 emerges as the optimal configuration, delivering the best overall results on both MultiEdit-Test and EmuEdit-Test. This validates the strategy of aggressively up-weighting the more complex G3 and G4 tasks. Thus, we adopt it as a standard setup of DMTL for our primary experiments.

\paragraph{Sample-wise Loss Re-weighting.}
To complement the DMTL strategy, we also investigate the loss-driven MTL (LMTL) paradigm by implementing the two sample-wise re-weighting strategies defined in Sec.~\ref{sec:dataset_MTL}. First, corresponding to Eq.~\ref{eq:lmtl_task_group}, we implement the heuristic intra-batch task group balancing using three pre-defined categories: (1) style transfer, (2) the most challenging text and GUI editing, and (3) all others. Here, we test a standard version ($\alpha=0$) with direct inverse frequency and a smoothed version applying additive smoothing with the smoothing factor $\alpha=1.0$ to the group counts, enhancing training stability. Second, we explore the more dynamic gradient-based sample re-weighting approach that re-weights each sample based on the $L_2$ norm of the gradient of the per-sample loss with respect to the model's noise prediction, as $G_i = \| \nabla_{\hat{\epsilon}_i} \ell_i \|_2$. Corresponding to Eq.~\ref{eq:lmtl_gradient_norm}, we test two opposing strategies: gradient focusing ($\omega_i \propto G_i$), which uses directly proportional weights to prioritize challenging examples, and gradient balancing ($\omega_i \propto G_i^{-1}$), which assigns weights inversely proportional to the gradient norm to improve stability.

As illustrated in the LMTL part of Tab.~\ref{tab:multitask}, for the intra-batch task group balancing loss strategies, the un-smoothed version slightly outperforms its smoothed counterpart. In gradient-based re-weighting strategies, gradient balancing employing inverse proportionality demonstrates superior performance to gradient focusing, and emerges as the best-performing method within the LMTL group. Nevertheless, this improvement is surpassed by the more direct and effective DMTL strategy of task-weighted data sampling, which achieves higher performance gains on MultiEdit-Test.

\subsection{Mixed Data Training}
To validate that mixing our challenging MultiEdit-Train data with broader external data enhances general task compatibility, we fine-tune SD3 on our MultiEdit-Train set while incrementally mixing UltraEdit data from 50K to 200K samples, with detailed results in Tab.~\ref{tab:sd3_mix}. Compared to SFT solely on the MultiEdit-Train set, this mixed data strategy yields marginal improvements on MultiEdit-Test but delivers more significant gains on EmuEdit-Test, achieving peak scores of 0.8409 on CLIP\textsubscript{img} and 0.7668 on DINO. This result suggests a strong synergy between the complex tasks in MultiEdit and the more foundational editing tasks from UltraEdit. Based on these results, we adopt the 100K data mixing configuration for our main method comparisons, as presented in Tab.~\ref{tab:quantitative_results}.

\section{Related Work}
\paragraph{Instruction-based Image Editing Models.}
Instruction-based image editing (IBIE) is a rapidly advancing field that enables image modification through natural language commands. InstructPix2Pix~\citep{Instructpix2pix} pioneers this paradigm by a prompt-to-prompt~\citep{hertz2022prompt} method and generating a large-scale synthetic dataset for IBIE. In contrast, MagicBrush~\citep{Magicbrush} demonstrates the value of real-world data by manually curating a high-quality yet expensive 10K-pair dataset from human annotators. Building on these foundational methods, subsequent research introduces numerous advancements. Some models focus on boosting multi-modal interaction and instruction fidelity through novel architectural designs, including SmartEdit~\citep{huang2024smartedit}, UltraEdit~\citep{zhao2024ultraedit}, and AnyEdit~\citep{yu2024anyedit}. In parallel, other works concentrate on training efficiency. For example, ICEdit~\citep{zhang2025context} employs LoRA-MoE hybrid tuning to achieve strong performance while minimizing training costs. Concurrently, another prominent line of research unlocks new possibilities by integrating MLLM-based multi-modal reasoning with diffusion-based controllable generation, like Step1X-Edit~\citep{liu2025step1x} and Qwen-Image~\citep{wu2025qwen}.

\paragraph{Instruction-based Image Editing Datasets.} 
High-quality training data are the cornerstone for developing powerful IBIE models. While foundational datasets like InstructPix2Pix and MagicBrush pave the way, subsequent research largely focuses on increasing scale and generation fidelity. For instance, Emu Edit~\citep{Emuedit} scales its dataset to 10 million proprietary pairs but offers limited public access. HQ-Edit~\citep{hui2024hq} uses DALL-E 3 to generate 197K high-fidelity pairs, though these pairs lack the pixel-to-pixel alignment critical for fine-grained details. In a different approach, UltraEdit~\citep{zhao2024ultraedit} contributes 4 million pairs using an inpainting-based strategy. Beyond scale and fidelity, the diversity and complexity of task design are equally critical. Datasets like AnyEdit~\citep{yu2024anyedit} begin to address this by curating 2.5 million pairs across a wide array of tasks, including more complex types like viewpoint and implicit editing. However, existing datasets still lack the breadth to cover multifaceted real-world scenarios and the depth required for training models on complex compositional and detail-oriented editing tasks.

\paragraph{The SOTA ImageGen-related Synthetic Datasets}
The recent release of the SOTA ImageGen (i.e., GPT-Image-1) marks a significant milestone in multi-modal generative modeling. Following its introduction, several empirical studies benchmark its capabilities, evaluating a heterogeneous set of tasks from generation quality and editing proficiency to advanced multi-modal reasoning~\citep{chen2025empirical, yan2025gpt, cao2025preliminary}. Capitalizing on these advanced capabilities, a key research trend is to leverage the SOTA ImageGen to generate curated training corpora, in an effort to bridge the performance gap between open-source and proprietary models. For instance, OmniConsistency~\citep{song2025omniconsistency} utilizes the SOTA ImageGen's exceptional stylization consistency to generate 2.6K style transfer image pairs. Focusing on more complex generation, Echo-4o-Image~\citep{ye2025echo4o} provides an approximately 180K-sample dataset generated by the SOTA ImageGen to boost model performance on demanding tasks like multi-reference synthesis and complex instruction execution. Similarly, ShareGPT-4o-Image~\citep{chen2025sharegpt} presents a 91K dataset, divided into 45K T2I samples and 46K IBIE examples. At the most enormous scale, GPT-Image-Edit-1.5M~\citep{wang2025gpt} constructs a massive IBIE corpus of over 1.5 million samples by unifying and refining existing datasets.

Following this line of work, based on the SOTA ImageGen, we introduce MultiEdit, a new large-scale IBIE dataset specifically targeting a spectrum of diverse and challenging image editing tasks. Structurally, this dataset comprises 6 distinct categories, spanning 18 non-style-transfer editing types and a comprehensive style transfer component with 38 styles across 3 groups.

\section{Conclusion}
In this work, we introduce MultiEdit, a comprehensive dataset for advancing instruction-based image editing beyond simple scenarios. It features over 107K high-quality editing samples from 6 challenging task categories, spanning 18 non-style-transfer editing types and a comprehensive style transfer component with 38 styles. Our dataset pipeline leverages the SOTA MLLM for visual-adaptive instruction generation and the SOTA ImageGen for high-fidelity edited image generation. We formally partition the complete collection into a large-scale MultiEdit-Train set ($\sim$106K samples) for model training and a carefully curated MultiEdit-Test benchmark (1.1K samples) for rigorous evaluation. Our experiments demonstrate that fine-tuning foundational models (SD3-based) with our MultiEdit-Train set, along with explored multi-task learning strategies, greately improves their sophisticated editing capabilities while effectively preserving performance on the standard benchmark. Ultimately, we believe MultiEdit provides a crucial resource to facilitate exploration into more diverse and challenging capabilities.

\section*{Limitations}
While MultiEdit represents an advance in diverse and challenging IBIE tasks, its current limitations also delineate clear pathways for our future research. First, as a supplementary dataset targeting diverse and challenging image editing tasks, MultiEdit requires integration with broader foundational corpora to train a general-purpose IBIE model from scratch effectively. Second, in experiments, the full potential of our dataset is constrained by the current SD-based architecture. Thus, a critical next step involves leveraging next-generation models, such as FLUX and unified multi-modal architectures, to unlock higher performance levels on complex image editing tasks.

%%%%%%%%%%%%%%%%%%%%%%%%%%%%%%%%%%%%%%%%%%%%%%%%%%%%%%%%%%%%%%%%%%
%%%%%%%%%%%%%%%%%%%%%%%%%%%%%%%%%%%%%%%%%%%%%%%%%%%%%%%%%%%%%%%%%%
%%%%%%%%%%%%%%%%%%%%%%%%%%%%%%%%%%%%%%%%%%%%%%%%%%%%%%%%%%%%%%%%%

\bibliographystyle{antgroup}

% \bibliography{reference}
\bibliography{main}

\clearpage
\appendix
\newtcolorbox{mydisplaybox_1}[1][]{
  breakable,
  enhanced, 
  arc=3mm,
  boxrule=0.5pt,
  colback=myBlueBase!5!white,         
  colframe=myBlueBase!80!black,       
  fonttitle=\bfseries,
  coltitle=white,               
  attach boxed title to top left={xshift=2.5mm, yshift=-2mm},
  boxed title style={
    arc=3mm,
    boxrule=0.3pt,              
    colback=myBlueBase!90!white,
    colframe=myBlueBase!80!black, 
  },
  drop shadow={black!20!white},
  #1 
}

\section{Text Editing Related to Chinese Characters} 

\begin{figure*}[!t]
\centering
\includegraphics[width=0.85\linewidth]{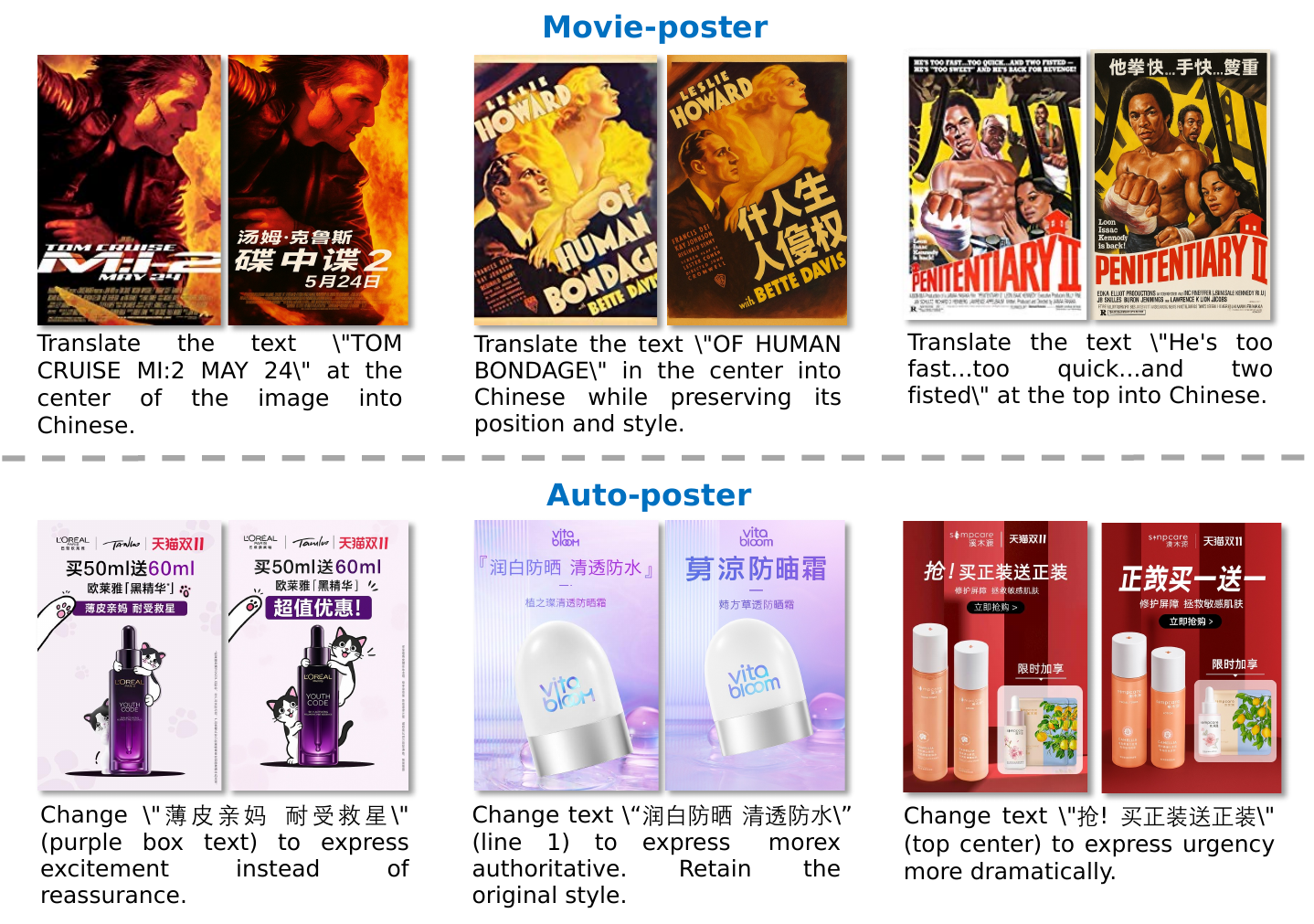}
\caption{Editing triplets involving Chinese elements, sampled from the Movie-poster and Auto-poster datasets. Each row contrasts one successful edit (the first triplet) with two failure cases, revealing the limitations of the SOTA ImageGen's Chinese text handling abilities.}
\label{fig:ana_chinese}
\end{figure*}

Given the growing attention on rendering non-alphabetic languages~\citep{wu2025qwen}, we evaluate the SOTA ImageGen's Chinese text handling abilities during the construction of MultiEdit. As shown in Fig.~\ref{fig:ana_chinese}, for the text editing task based on the Movie-poster dataset~\citep{dewidar2019inferring}, the model demonstrates reasonable performance in translating short English phrases into Chinese. However, rendering longer Chinese texts leads to diminished stability and detail preservation, often failing to achieve flawless presentation. The challenges are even more pronounced with the Auto-poster dataset~\citep{lin2023autoposter} for text editing, which contains substantial pre-existing Chinese advertising text. Corruptions and inaccuracies are commonly observed when using the SOTA ImageGen model to edit in-image text, underscoring the current challenges in this text editing domain. This observation of the SOTA ImageGen aligns with the findings in recent work~\citep{yan2025gpt,song2025omniconsistency,wu2025qwen}.

Consequently, due to these identified shortcomings, we exclude the Auto-poster subset for text editing and the Chinese Translation editing type of related tasks from the final MultiEdit dataset to ensure its overall quality and effectiveness. We leave the challenge of high-quality image editing involving non-alphabetic languages, particularly Chinese, for our future work.

\section{Systematic Expert-designed Meta-instructions}\label{sec:app_meta_instruc}
As detailed in Sec.~\ref{sec:dataset_colloct}, our pipeline employs the SOTA MLLM to generate the visual-adaptive editing instruction from each source image. This generation process is structured by a set of expert-designed meta-instructions. For style transfer, we utilize a template-based strategy to encompass diverse artistic styles. For other complex tasks, we systematically define a suite of meta-instructions, which are enumerated below.

\begin{mydisplaybox_1}[title={Object Reference Editing}]
\begin{description}
\item[\textbf{(1) Color:}] Please randomly pick an object (other than a person) from the image, and give an instruction for the image editing model to change its \textbf{color}. Please describe the picked object's position precisely with words (e.g., 2nd chair from the left), and output the edit instruction directly without any additional words. Please make sure the edit instruction is within 50 tokens.

\item[\textbf{(2) Shape:}] Please randomly pick an object (other than a person) from the image, and give an instruction for the image editing model to change its \textbf{shape}. Please describe the picked object's position precisely with words (e.g., 2nd chair from the left), and output the edit instruction directly without any additional words. Please make sure the edit instruction is within 50 tokens.

\item[\textbf{(3) Scale:}] Please randomly pick an object (other than a person) from the image, and give an instruction for the image editing model to change its \textbf{scale}. Please describe the picked object's position precisely with words (e.g., 2nd chair from the left), and output the edit instruction directly without any additional words. Please make sure the edit instruction is within 50 tokens.

\item[\textbf{(4) Position:}] Please randomly pick an object (other than a person) from the image, and give an instruction for the image editing model to change its \textbf{position}. Please describe the picked object's position precisely with words (e.g., 2nd chair from the left), and output the edit instruction directly without any additional words. Please make sure the edit instruction is within 50 tokens.
\end{description}
\end{mydisplaybox_1}

\begin{mydisplaybox_1}[title={Person Reference Editing}]
\begin{description}
\item[\textbf{(1) Pose:}] Please randomly pick a person from the image, and give an instruction for the image editing model to change its \textbf{pose}. Please describe the picked person's position precisely with words (e.g., 2nd man from the left), and output the edit instruction directly without any additional words. Please make sure the edit instruction is within 50 tokens.

\item[\textbf{(2) Clothing:}] Please randomly pick a person from the image, and give an instruction for the image editing model to change its \textbf{clothing}. Please describe the picked person's position precisely with words (e.g., 2nd man from the left), and output the edit instruction directly without any additional words. Please make sure the edit instruction is within 50 tokens. 

\item[\textbf{(3) Hairstyle:}] Please randomly pick a person from the image, and give an instruction for the image editing model to change its \textbf{hairstyle}. Please describe the picked person's position precisely with words (e.g., 2nd man from the left), and output the edit instruction directly without any additional words. Please make sure the edit instruction is within 50 tokens.

\item[\textbf{(4) Skin Color:}] Please randomly pick a person from the image, and give an instruction for the image editing model to change its \textbf{skin color}. Please describe the picked person's position precisely with words (e.g., 2nd man from the left), and output the edit instruction directly without any additional words. Please make sure the edit instruction is within 50 tokens.

\item[\textbf{(5) Figure:}] Please randomly pick a person from the image, and give an instruction for the image editing model to change its \textbf{figure}. Please describe the picked person's position precisely with words (e.g., 2nd man from the left), and output the edit instruction directly without any additional words. Please make sure the edit instruction is within 50 tokens.
\end{description}
\end{mydisplaybox_1}

\begin{mydisplaybox_1}[title={Text Editing}]
\begin{description}
\item[\textbf{(1) Font Style:}] Please randomly select some text or a paragraph from the image, and provide an instruction for the image editing model to change its \textbf{font style}. Precisely indicate the position of the selected content (e.g., line 2-3, 2nd paragraph, "use this on-line proclamation"), and output the edit instruction directly without any additional words. Ensure the instruction is within 50 tokens, and avoid selecting too small pieces of content.

\item[\textbf{(2) Expression:}] Please randomly select some text or a paragraph from the image, and provide an instruction for the image editing model to change its \textbf{expression}. Precisely indicate the position of the selected content (e.g., line 2-3, 2nd paragraph, "use this on-line proclamation"), and output the edit instruction directly without any additional words. Ensure the instruction is within 50 tokens, and avoid selecting too small pieces of content.

\item[\textbf{(3) Display Medium:}] Please randomly provide an instruction for the image editing model to change the given document's \textbf{display medium} (e.g., computer screen and notebook). Please output the edit instruction directly without any additional words. Ensure the instruction is within 50 tokens.

\item[\textbf{(4) Font Color:}] Please randomly select some text or a paragraph from the image, and provide an instruction for the image editing model to change its \textbf{font color}. Precisely indicate the position of the selected content (e.g., line 2-3, 2nd paragraph, "use this on-line proclamation"), and output the edit instruction directly without any additional words. Ensure the instruction is within 50 tokens, and avoid selecting too small pieces of content.
\end{description}
\end{mydisplaybox_1}

\begin{mydisplaybox_1}[title={GUI Editing}]
\begin{description}
\item[\textbf{(1) Icon Attributes:}] Please randomly pick an \textbf{icon} from the image, and give an instruction for the image editing model to change its \textbf{attributes} (e.g., color and shape). Please describe the picked icon position precisely with words (e.g., the top right home icon), and output the edit instruction directly without any additional words. Please make sure the edit instruction is within 50 tokens.

\item[\textbf{(2) Display Medium:}] Please randomly provide an instruction for the image editing model to change the given GUI's \textbf{display medium} (e.g., computer screen and notebook). Please output the edit instruction directly without any additional words. Ensure the instruction is within 50 tokens.
\end{description}
\end{mydisplaybox_1}

\begin{mydisplaybox_1}[title={View Editing}]
\begin{description}
\item[\textbf{(1) Person:}] Based on this image, create an editing instruction for the image editing model to generate alternative views of the \textbf{person} shown. Please provide the instruction directly without additional commentary, and keep it within 50 tokens. Ensure the new perspective maintains some overlap with the original image.

\item[\textbf{(2) Landmark:}] Based on this image, create an editing instruction for the image editing model to generate alternative views of the \textbf{landmark} shown. Please provide the instruction directly without additional commentary, and keep it within 50 tokens. Ensure the new perspective maintains some overlap with the original image.

\item[\textbf{(3) Object:}] Based on this image, create an editing instruction for the image editing model to generate alternative views of the \textbf{object} shown. Please provide the instruction directly without additional commentary, and keep it within 50 tokens. Ensure the new perspective maintains some overlap with the original image.
\end{description}
\end{mydisplaybox_1}

\section{Model Inference Guidance Experiments}\label{sec:app_guidance}
For model inference, we employ a consistent setting of 50 inference steps for all models, with the exception of AnyEdit, for which we use its default of 100 steps. Regarding the guidance\_scale, we adhere to the default configurations for baseline models. AnyEdit uses image\_guidance\_scale of 3.0 and text\_guidance\_scale of 3.0, and Step1X-Edit operates with a cfg\_guidance of 6.0. The foundational SD3 directly utilizes the hyperparameters of StableDiffusion3Img2ImgPipeline with a strength of 0.6 and a guidance\_scale of 7.0. 

We utilize the StableDiffusion3InstructPix2PixPipeline for the base UltraEdit and all our fine-tuned models. To determine the optimal guidance, we conduct detailed hyperparameter experiments for guidance\_scale based on UltraEdit and our enhanced ME-UEdit-DMTL, with results presented in Tab.~\ref{tab:guidance_test}. Based on this analysis and our qualitative observations, we adopt a universal text\_guidance\_scale of 4.0 for UltraEdit and our fine-tuned models. For the image\_guidance\_scale, UltraEdit sets it to 2.0. Our UltraEdit-based fine-tuned models use 3.5 on MultiEdit-Test and 5.0 on EmuEdit-Test, while our SD3-based variants maintain 3.5 on MultiEdit-Test and adopt 2.0 on EmuEdit-Test, aligning with the base UltraEdit's setting. The final parameter configurations for all models are summarized in Tab.~\ref{tab:guidance_config}.

\section{Detailed Quantitative Results on MultiEdit-Test}\label{sec:app_detail_multitest}
For a detailed task-level analysis, we present quantitative results in Tabs.~\ref{tab:detail_ref_gui} and~\ref{tab:detail_text_view_style} comparing the best-performing fine-tuned method based on our MultiEdit-Train set (ME-UEdit-DMTL) with four baseline methods, which are also supported by the qualitative examples in Figs.~\ref{fig:qualita_1} and~\ref{fig:qualita_2}. Specifically, ME-UEdit-DMTL consistently excels across a diverse range of challenging tasks, including object reference editing, GUI editing, text editing, and style transfer. In contrast, baseline SOTA methods show apparent limitations in specific scenarios. For instance, Step1X-Edit struggles with challenging text editing tasks involving complex display mediums, while AnyEdit's effectiveness is restricted in style transfer scenarios.

\begin{table*}[!tb]
\centering
\caption{Guidance scale tests on MultiEdit-Test and EmuEdit-Test. Best results per group are shown in \textbf{bold}.}
\label{tab_guidance_combined}
\renewcommand{\arraystretch}{1.25}
\setlength{\tabcolsep}{3pt}
\resizebox{\textwidth}{!}{
\begin{tabular}{lc ccccc cccccc}
\toprule
\multirow{2}{*}{\textbf{Method}} & \multirow{2}{*}{\textbf{Text-G}} & \multicolumn{5}{c}{\textbf{MultiEdit-Test}} & \multicolumn{6}{c}{\textbf{EmuEdit-Test}} \\
\cmidrule(lr){3-7} \cmidrule(lr){8-13}
& & \textbf{Image-G} & CLIP\textsubscript{img}$\uparrow$ & DINO$\uparrow$ & L\textsubscript{1}$\downarrow$ & L\textsubscript{2}$\downarrow$ 
& \textbf{Image-G} & CLIP\textsubscript{img}$\uparrow$ & CLIP\textsubscript{txt}$\uparrow$ & DINO$\uparrow$ & L\textsubscript{1}$\downarrow$ & L\textsubscript{2}$\downarrow$ \\
\midrule
% --- UltraEdit Block ---
\multirow{12}{*}{\textbf{UltraEdit}} 
& \multirow{4}{*}{5.0} 
& 1.5 & 0.7826 & 0.7072 & 0.1938 & 0.0846 & 1.5 & 0.8538 & \bfseries 0.2741 & 0.7688 & 0.1197 & 0.0421 \\
& & 2.0 & 0.7932 & 0.7229 & 0.1994 & 0.0882 & 2.0 & 0.8649 & 0.2722 & 0.7936 & 0.1177 & 0.0419 \\
& & 2.5 & 0.7943 & 0.7198 & 0.2067 & 0.0927 & 2.5 & 0.8643 & 0.2703 & 0.7952 & 0.1219 & 0.0436 \\
& & 3.0 & 0.7907 & 0.7107 & 0.2137 & 0.0974 & 3.0 & 0.8566 & 0.2684 & 0.7850 & 0.1297 & 0.0466 \\
\cmidrule(lr){2-13}
& \multirow{4}{*}{4.5} 
& 1.5 & 0.7882 & 0.7163 & 0.1925 & 0.0839 & 1.5 & 0.8580 & 0.2740 & 0.7773 & 0.1173 & 0.0414 \\
& & 2.0 & 0.7968 & 0.7264 & 0.1988 & 0.0878 & 2.0 & 0.8662 & 0.2708 & 0.7969 & 0.1159 & 0.0412 \\
& & 2.5 & 0.7983 & 0.7235 & 0.2056 & 0.0921 & 2.5 & 0.8647 & 0.2695 & 0.7986 & 0.1207 & 0.0432 \\
& & 3.0 & 0.7920 & 0.7133 & 0.2129 & 0.0968 & 3.0 & 0.8568 & 0.2679 & 0.7880 & 0.1285 & 0.0461 \\
\cmidrule(lr){2-13}
& \multirow{4}{*}{4.0} 
& 1.5 & 0.7931 & 0.7210 & \bfseries 0.1919 & \bfseries 0.0835 & 1.5 & 0.8611 & 0.2725 & 0.7862 & 0.1153 & 0.0406 \\
& & 2.0 & \bfseries 0.8017 & \bfseries 0.7303 & 0.1976 & 0.0871 & 2.0 & \bfseries 0.8685 & 0.2706 & \bfseries 0.8055 & \bfseries 0.1137 & \bfseries 0.0406 \\
& & 2.5 & 0.7997 & 0.7283 & 0.2044 & 0.0913 & 2.5 & 0.8651 & 0.2682 & 0.8018 & 0.1191 & 0.0426 \\
& & 3.0 & 0.7934 & 0.7156 & 0.2118 & 0.0961 & 3.0 & 0.8580 & 0.2663 & 0.7905 & 0.1273 & 0.0457 \\
\midrule
% --- ME-UEdit-DMTL Block ---
\multirow{12}{*}{\shortstack{\textbf{ME-UEdit-DMTL}}}
& \multirow{4}{*}{5.0} 
& 3.0 & 0.8096 & 0.7985 & 0.1847 & 0.0833 & 4.5 & 0.8366 & \bfseries 0.2737 & 0.7555 & 0.1957 & 0.0834 \\
& & 3.5 & 0.8129 & 0.8024 & 0.1838 & 0.0818 & 5.0 & \bfseries 0.8383 & 0.2716 & 0.7570 & 0.1944 & 0.0824 \\
& & 4.0 & 0.8136 & 0.7984 & 0.1852 & 0.0826 & 5.5 & 0.8374 & 0.2703 & 0.7585 & 0.1928 & 0.0813 \\
& & 4.5 & 0.8148 & 0.7986 & 0.1862 & 0.0825 & 6.0 & 0.8378 & 0.2693 & 0.7596 & 0.1920 & 0.0808 \\
\cmidrule(lr){2-13}
& \multirow{4}{*}{4.5}
& 3.0 & 0.8122 & 0.8039 & 0.1828 & 0.0815 & 4.5 & 0.8376 & 0.2719 & 0.7578 & 0.1937 & 0.0817 \\
& & 3.5 & 0.8153 & 0.8028 & 0.1831 & 0.0813 & 5.0 & 0.8377 & 0.2704 & 0.7593 & 0.1922 & 0.0811 \\
& & 4.0 & 0.8161 & 0.8052 & 0.1832 & 0.0808 & 5.5 & 0.8378 & 0.2690 & 0.7604 & 0.1902 & 0.0796 \\
& & 4.5 & 0.8160 & 0.8000 & 0.1842 & 0.0807 & 6.0 & 0.8369 & 0.2673 & 0.7604 & 0.1899 & 0.0794 \\
\cmidrule(lr){2-13}
& \multirow{4}{*}{4.0}
& 3.0 & 0.8168 & 0.8069 & \bfseries 0.1795 & \bfseries 0.0791 & 4.5 & 0.8381 & 0.2696 & 0.7587 & 0.1910 & 0.0802 \\
& & 3.5 & \bfseries 0.8174 & \bfseries 0.8071 & 0.1814 & 0.0800 & 5.0 & 0.8376 & 0.2681 & \bfseries 0.7610 & 0.1896 & 0.0793 \\
& & 4.0 & 0.8171 & 0.8060 & 0.1812 & \bfseries 0.0791 & 5.5 & 0.8377 & 0.2672 & 0.7608 & 0.1898 & 0.0793 \\
& & 4.5 & 0.8162 & 0.7985 & 0.1857 & 0.0818 & 6.0 & 0.8382 & 0.2661 & 0.7591 & \bfseries 0.1892 & \bfseries 0.0789 \\
\bottomrule
\end{tabular}
}
\label{tab:guidance_test}
\end{table*}

\begin{table*}[!tb]
\centering
\caption{Inference hyperparameters summary for all methods.}
\renewcommand{\arraystretch}{1.25}
\setlength{\tabcolsep}{3pt}
\resizebox{\textwidth}{!}{
\begin{tabular}{lcc}
\toprule
\textbf{Method} & \textbf{Inference Step} & \textbf{Guidance Config} \\
\midrule
% --- Baseline Models Section ---
SD3 & 50 & strength = 0.6, guidance\_scale = 7.0  \\
UltraEdit & 50 & image\_guidance\_scale = 2.0, text\_guidance\_scale = 4.0 \\
AnyEdit & 100 & image\_guidance\_scale = 3.0, text\_guidance\_scale = 3.0 \\
Step1x-Edit & 50 & cfg\_guidance = 6.0 \\
\midrule
% --- Fine-tuned Models Section ---
\multirow{2}{*}{SFT models for MultiEdit-Test} & \multirow{2}{*}{50} & SD3-based: image\_guidance\_scale = 3.5, text\_guidance\_scale = 4.0  \\
& & UltraEdit-based: image\_guidance\_scale = 3.5, text\_guidance\_scale = 4.0  \\
\hline
\multirow{2}{*}{SFT models for EmuEdit-Test} & \multirow{2}{*}{50} & SD3-based: image\_guidance\_scale = 2.0, text\_guidance\_scale = 4.0  \\
& & UltraEdit-based: image\_guidance\_scale = 5.0, text\_guidance\_scale = 4.0  \\
\bottomrule
\end{tabular}
}
\label{tab:guidance_config}
\end{table*}

% Detailed breakdown on reference and GUI editing tasks ---
\begin{table*}[!t]
\centering
\caption{Task-level quantitative comparison of different methods on reference and GUI editing from our MultiEdit-Test benchmark. Best and second-best results are shown in \textbf{bold} and \underline{underlined}, respectively.}
\renewcommand{\arraystretch}{1.25}
\setlength{\tabcolsep}{3pt} 
\resizebox{\textwidth}{!}{
\begin{tabular}{l cccc ccccc cc} 
\toprule
% --- LEVEL 1 HEADERS ---
\multirow{3}{*}{\textbf{Metric}} & \multicolumn{4}{c}{\textbf{Object Reference Editing}} & \multicolumn{5}{c}{\textbf{Person Reference Editing}} & \multicolumn{2}{c}{\textbf{GUI Editing}} \\
\cmidrule(lr){2-5} \cmidrule(lr){6-10} \cmidrule(lr){11-12}

% --- LEVEL 2 HEADERS ---
& color & shape & scale & position & pose & clothing & hairstyle & skin & figure & icon & display \\
\midrule
\midrule

% --- BLOCK 1: SD3 ---
\multicolumn{12}{l}{\textbf{SD3}} \\
CLIP\textsubscript{img}$\uparrow$ & 0.7609 & 0.7135 & 0.7473 & 0.7608 & 0.6994 & 0.6757 & 0.6616 & 0.6855 & 0.7217 & 0.7033 & 0.6705 \\
DINO$\uparrow$                 & 0.6204 & 0.5311 & 0.5833 & 0.6201 & 0.6285 & 0.5864 & 0.5878 & 0.6308 & 0.6308 & 0.7457 & 0.7383 \\
L\textsubscript{1}$\downarrow$   & 0.1590 & 0.1572 & 0.1770 & 0.1675 & 0.1888 & 0.1723 & 0.1795 & 0.1708 & 0.1861 & 0.1337 & \underline{0.2700} \\
L\textsubscript{2}$\downarrow$   & 0.0542 & 0.0513 & 0.0636 & 0.0606 & 0.0710 & 0.0643 & 0.0659 & 0.0614 & 0.0724 & \bfseries 0.0690 & \underline{0.1724} \\
\midrule

% --- BLOCK 2: UltraEdit ---
\multicolumn{12}{l}{\textbf{UltraEdit}} \\
CLIP\textsubscript{img}$\uparrow$ & 0.8823 & 0.8278 & 0.8940 & 0.8851 & 0.8423 & 0.8419 & 0.8453 & 0.8582 & 0.8442 & \underline{0.7914} & \bfseries 0.7394 \\
DINO$\uparrow$                 & 0.8286 & 0.7402 & \underline{0.8289} & 0.8053 & 0.7610 & 0.7833 & 0.8138 & 0.8365 & 0.7721 & 0.8153 & \underline{0.7875} \\
L\textsubscript{1}$\downarrow$   & 0.1477 & 0.1502 & \underline{0.1602} & 0.1625 & 0.1786 & 0.1688 & 0.1649 & 0.1630 & 0.1802 & 0.1324 & 0.2774 \\
L\textsubscript{2}$\downarrow$   & \underline{0.0493} & \underline{0.0491} & \underline{0.0565} & 0.0589 & 0.0680 & 0.0646 & 0.0608 & 0.0591 & 0.0708 & 0.0715 & 0.1811 \\
\midrule

% --- BLOCK FOR AnyEdit ---
\multicolumn{12}{l}{\textbf{AnyEdit}} \\
CLIP\textsubscript{img}$\uparrow$ & \bfseries 0.8935 & \underline{0.8600} & \underline{0.8971} & 0.8626 & \underline{0.8506} & \underline{0.8695} & \underline{0.8515} & \underline{0.8726} & \underline{0.8484} & 0.7815 & 0.7076 \\
DINO$\uparrow$                 & \underline{0.8469} & \underline{0.7868} & 0.8247 & 0.7632 & 0.7598 & 0.8149 & 0.8238 & \bfseries 0.8591 & 0.7727 & 0.7817 & 0.6854 \\
L\textsubscript{1}$\downarrow$   & \bfseries 0.1414 & \bfseries 0.1392 & 0.1660 & \underline{0.1559} & \underline{0.1713} & \underline{0.1561} & \bfseries 0.1578 & \bfseries 0.1523 & \bfseries 0.1696 & 0.1582 & 0.3278 \\
L\textsubscript{2}$\downarrow$   & \bfseries 0.0475 & \bfseries 0.0441 & 0.0595 & \underline{0.0559} & \bfseries 0.0628 & \bfseries 0.0585 & \bfseries 0.0558 & \bfseries 0.0526 & \bfseries 0.0649 & 0.0847 & 0.2009 \\
\midrule

% --- BLOCK FOR step1x-edit ---
\multicolumn{12}{l}{\textbf{Step1X-Edit}} \\
CLIP\textsubscript{img}$\uparrow$ & \underline{0.8867} & 0.8549 & \bfseries 0.9010 & \bfseries 0.8947 & \bfseries 0.8626 & \bfseries 0.8812 & \bfseries 0.8682 & \bfseries 0.8828 & \bfseries 0.8619 & \bfseries 0.8872 & 0.6037 \\
DINO$\uparrow$                 & 0.8075 & 0.7329 & 0.8250 & \underline{0.8263} & \underline{0.7748} & \bfseries 0.8319 & \underline{0.8464} & 0.8430 & \underline{0.7953} & \bfseries 0.8951 & 0.4499 \\
L\textsubscript{1}$\downarrow$   & 0.1473 & 0.1523 & 0.1647 & \bfseries 0.1532 & 0.1789 & 0.1572 & \underline{0.1596} & 0.1569 & \underline{0.1707} & \bfseries 0.1271 & 0.3723 \\
L\textsubscript{2}$\downarrow$   & 0.0529 & 0.0529 & 0.0601 & \bfseries 0.0546 & \underline{0.0675} & \underline{0.0593} & \underline{0.0568} & \underline{0.0556} & \underline{0.0650} & \underline{0.0696} & 0.2335 \\
\midrule

% --- BLOCK FOR UltraEdit+MultiEdit ---
\multicolumn{12}{l}{\textbf{ME-UEdit-DMTL}} \\
CLIP\textsubscript{img}$\uparrow$ & 0.8809 & \bfseries 0.8607 & 0.8912 & \underline{0.8905} & 0.8334 & 0.8546 & 0.8401 & 0.8551 & 0.8467 & 0.7744 & \underline{0.7220} \\
DINO$\uparrow$                 & \bfseries 0.8473 & \bfseries 0.8041 & \bfseries 0.8419 & \bfseries 0.8418 & \bfseries 0.8050 & \underline{0.8304} & \bfseries 0.8510 & \underline{0.8570} & \bfseries 0.8147 & \underline{0.8448} & \bfseries 0.8047 \\
L\textsubscript{1}$\downarrow$   & 0.1470 & \underline{0.1435} & \bfseries 0.1568 & 0.1593 & \bfseries 0.1708 & \bfseries 0.1545 & 0.1633 & \underline{0.1545} & 0.1711 & \underline{0.1311} & \bfseries 0.2378 \\
L\textsubscript{2}$\downarrow$   & 0.0528 & 0.0490 & \bfseries 0.0564 & 0.0601 & 0.0677 & 0.0596 & 0.0622 & 0.0581 & 0.0690 & 0.0710 & \bfseries 0.1485 \\

\bottomrule
\end{tabular}
\label{tab:detail_ref_gui}
}
\end{table*}

% Detailed breakdown on text, view editing, and style transfer tasks ---
\begin{table*}[!t]
\centering
\caption{Task-level quantitative comparison of different methods on text editing, view editing, and style transfer sub-tasks from our MultiEdit-Test benchmark. Best and second-best results are shown in \textbf{bold} and \underline{underlined}, respectively.}
\renewcommand{\arraystretch}{1.25}
\setlength{\tabcolsep}{3pt} 
\resizebox{\textwidth}{!}{
\begin{tabular}{l cccc ccc cccc} 
\toprule
% --- LEVEL 1 HEADERS ---
\multirow{2}{*}{\textbf{Metric}} & \multicolumn{4}{c}{\textbf{Text Editing}} & \multicolumn{3}{c}{\textbf{View Editing}} & \multicolumn{4}{c}{\textbf{Style Transfer}} \\
\cmidrule(lr){2-5} \cmidrule(lr){6-8} \cmidrule(lr){9-12}

% --- LEVEL 2 HEADERS ---
& fontStyle & expression & display &fontColor & person & landmark & object & person & animal & landmark & object \\
\midrule

% --- BLOCK 1: SD3 ---
\multicolumn{12}{l}{\textbf{SD3}} \\
CLIP\textsubscript{img}$\uparrow$ & 0.6375 & 0.6202 & 0.4809 & 0.6089 & 0.6782 & 0.7093 & 0.8136 & 0.7087 & 0.6584 & 0.6313 & 0.6505 \\
DINO$\uparrow$                 & 0.5245 & 0.5183 & 0.4423 & 0.5300 & 0.7026 & 0.6124 & 0.6673 & 0.5619 & 0.3642 & 0.4319 & 0.3927 \\
L\textsubscript{1}$\downarrow$   & 0.2188 & 0.2063 & 0.3374 & 0.2148 & \underline{0.1327} & 0.1978 & 0.1880 & 0.1933 & 0.2406 & 0.2452 & 0.2266 \\
L\textsubscript{2}$\downarrow$   & 0.1062 & 0.0975 & 0.1894 & 0.1093 & \underline{0.0504} & \underline{0.0773} & 0.0699 & 0.0836 & 0.0986 & 0.1031 & 0.0878 \\
\midrule

% --- BLOCK 2: UltraEdit ---
\multicolumn{12}{l}{\textbf{UltraEdit}} \\
CLIP\textsubscript{img}$\uparrow$ & 0.7368 & 0.7324 & \bfseries 0.7080 & 0.7706 & 0.8395 & 0.8215 & 0.8738 & 0.7107 & 0.7458 & 0.7052 & 0.7410 \\
DINO$\uparrow$                 & 0.6774 & 0.6917 & \underline{0.6900} & 0.7204 & 0.8184 & 0.7199 & 0.7641 & 0.6280 & 0.4510 & 0.5424 & 0.5897 \\
L\textsubscript{1}$\downarrow$   & 0.2175 & 0.2066 & 0.3464 & 0.2132 & 0.1458 & 0.2017 & 0.1867 & 0.2070 & 0.2566 & 0.2493 & 0.2315 \\
L\textsubscript{2}$\downarrow$   & 0.1071 & 0.0996 & 0.1984 & 0.1108 & 0.0578 & 0.0818 & 0.0711 & 0.0935 & 0.1101 & 0.1055 & 0.0911 \\
\midrule

% --- BLOCK FOR AnyEdit ---
\multicolumn{12}{l}{\textbf{AnyEdit}} \\
CLIP\textsubscript{img}$\uparrow$ & \underline{0.7556} & \underline{0.7392} & \underline{0.6528} & \underline{0.7816} & \underline{0.8507} & \bfseries 0.8435 & \underline{0.8869} & 0.7023 & 0.7741 & 0.7117 & 0.7462 \\
DINO$\uparrow$                 & 0.6852 & 0.6931 & 0.5802 & 0.7235 & 0.7925 & \underline{0.7460} & \bfseries 0.8023 & 0.5857 & 0.4545 & 0.5526 & 0.6093 \\
L\textsubscript{1}$\downarrow$   & 0.2113 & \underline{0.2011} & 0.3256 & \underline{0.2028} & 0.1912 & \bfseries 0.1904 & \bfseries 0.1712 & 0.1866 & \underline{0.2291} & \underline{0.2319} & \underline{0.2140} \\
L\textsubscript{2}$\downarrow$   & 0.1020 & \bfseries 0.0958 & 0.1741 & \underline{0.1003} & 0.0912 & \bfseries 0.0733 & \bfseries 0.0614 & \underline{0.0793} & \underline{0.0911} & \underline{0.0947} & \bfseries 0.0798 \\
\midrule

% --- BLOCK FOR step1x-edit ---
\multicolumn{12}{l}{\textbf{Step1X-Edit}} \\
CLIP\textsubscript{img}$\uparrow$ & \bfseries 0.7766 & \bfseries 0.7811 & 0.5886 & \bfseries 0.8033 & \bfseries 0.9190 & \underline{0.8393} & \bfseries 0.8991 & 0.8335 & 0.8626 & 0.8065 & 0.8434 \\
DINO$\uparrow$                 & 0.7101 & 0.6892 & 0.4585 & \underline{0.7555} & \bfseries 0.8633 & 0.7156 & 0.7797 & 0.7351 & 0.6714 & 0.6881 & 0.7304 \\
L\textsubscript{1}$\downarrow$   & \underline{0.2090} & 0.2040 & 0.3283 & 0.2042 & \bfseries 0.1242 & 0.2038 & \underline{0.1807} & 0.1967 & 0.2372 & 0.2440 & 0.2272 \\
L\textsubscript{2}$\downarrow$   & \underline{0.0994} & \underline{0.0962} & 0.1757 & 0.1021 & \bfseries 0.0486 & 0.0842 & \underline{0.0674} & 0.0838 & 0.0962 & 0.1025 & 0.0909 \\
\midrule

% --- BLOCK FOR UltraEdit+MultiEdit ---
\multicolumn{12}{l}{\textbf{ME-UEdit-DMTL}} \\
CLIP\textsubscript{img}$\uparrow$ & 0.7037 & 0.7016 & 0.6276 & 0.7187 & 0.8361 & 0.8170 & 0.8809 & \bfseries 0.8617 & \bfseries 0.8865 & \bfseries 0.8263 & \bfseries 0.8736 \\
DINO$\uparrow$                 & \bfseries 0.7736 & \bfseries 0.7758 & \bfseries 0.7093 & \bfseries 0.7881 & \underline{0.8547} & \bfseries 0.7686 & \underline{0.7863} & \bfseries 0.8226 & \bfseries 0.7880 & \bfseries 0.7531 & \bfseries 0.7938 \\
L\textsubscript{1}$\downarrow$   & \bfseries 0.1981 & \bfseries 0.1950 & \bfseries 0.2810 & \bfseries 0.1899 & 0.1581 & \underline{0.1952} & 0.1864 & \bfseries 0.1658 & \bfseries 0.2096 & \bfseries 0.2156 & \bfseries 0.2069 \\
L\textsubscript{2}$\downarrow$   & \bfseries 0.0989 & 0.0980 & \bfseries 0.1528 & \bfseries 0.0993 & 0.0783 & 0.0793 & 0.0740 & \bfseries 0.0682 & \bfseries 0.0855 & \bfseries 0.0880 & \underline{0.0829} \\

\bottomrule
\end{tabular}
}
\label{tab:detail_text_view_style}
\end{table*}

\end{document}